\documentclass[]{beingbeyond}
\usepackage{enumitem}
\usepackage[toc,page,header]{appendix}

\usepackage[utf8]{inputenc} 
\usepackage[T1]{fontenc}    
\usepackage{hyperref}       
\usepackage{url}            
\usepackage{array}          
\usepackage{booktabs}       
\usepackage{amsfonts}       
\usepackage{nicefrac}       
\usepackage{diagbox}  
\usepackage{microtype}      
\usepackage{xcolor}         
\usepackage{xspace}
\usepackage{bm}
\usepackage{bbding}
\usepackage{bbm}
\usepackage{tabularx}
\usepackage{amssymb}
\usepackage{enumitem}
\usepackage{amsmath}
\usepackage{mathtools}
\usepackage{amsthm}
\usepackage{multirow}
\usepackage{makecell}
\usepackage{color}
\usepackage{colortbl}
\usepackage{adjustbox}
\usepackage{caption}
\usepackage{graphicx}
\usepackage{soul} 
\usepackage{pifont}
\usepackage{wrapfig}
\usepackage{multicol}

\usepackage[table]{xcolor}

\usepackage{xspace}
\makeatletter
\DeclareRobustCommand\onedot{\futurelet\@let@token\@onedot}
\def\@onedot{\ifx\@let@token.\else.\null\fi\xspace}

\definecolor{BlockC}{gray}{0.98}  
\definecolor{BlockA}{RGB}{191,211,230}
\definecolor{BlockB}{RGB}{199,233,192} 
\newcommand{\benchmark}{BeTTER\ }

\definecolor{cgreen}{RGB}{0, 150, 0}
\definecolor{cred}{RGB}{200, 0, 0}
\definecolor{cblue}{RGB}{0, 0, 150}

\title{Unmasking the Illusion of Embodied Reasoning in Vision-Language-Action Models}

\author{{\bfseries
Haiweng Xu$^{1,3}$~
Sipeng Zheng$^{3}$~
Hao Luo$^{1,3}$~
Wanpeng Zhang$^{1,3}$ \\
Ziheng Xi$^{2,3}$~
Zongqing Lu$^{1,3,\dagger}$
}}

\affiliation{
$^{1}$Peking University \quad
$^{2}$Tsinghua University \quad
$^{3}$BeingBeyond \quad
}

\webpage{\url{https://research.beingbeyond.com/better}}

\firstfig[width=\linewidth][\textwidth]
  {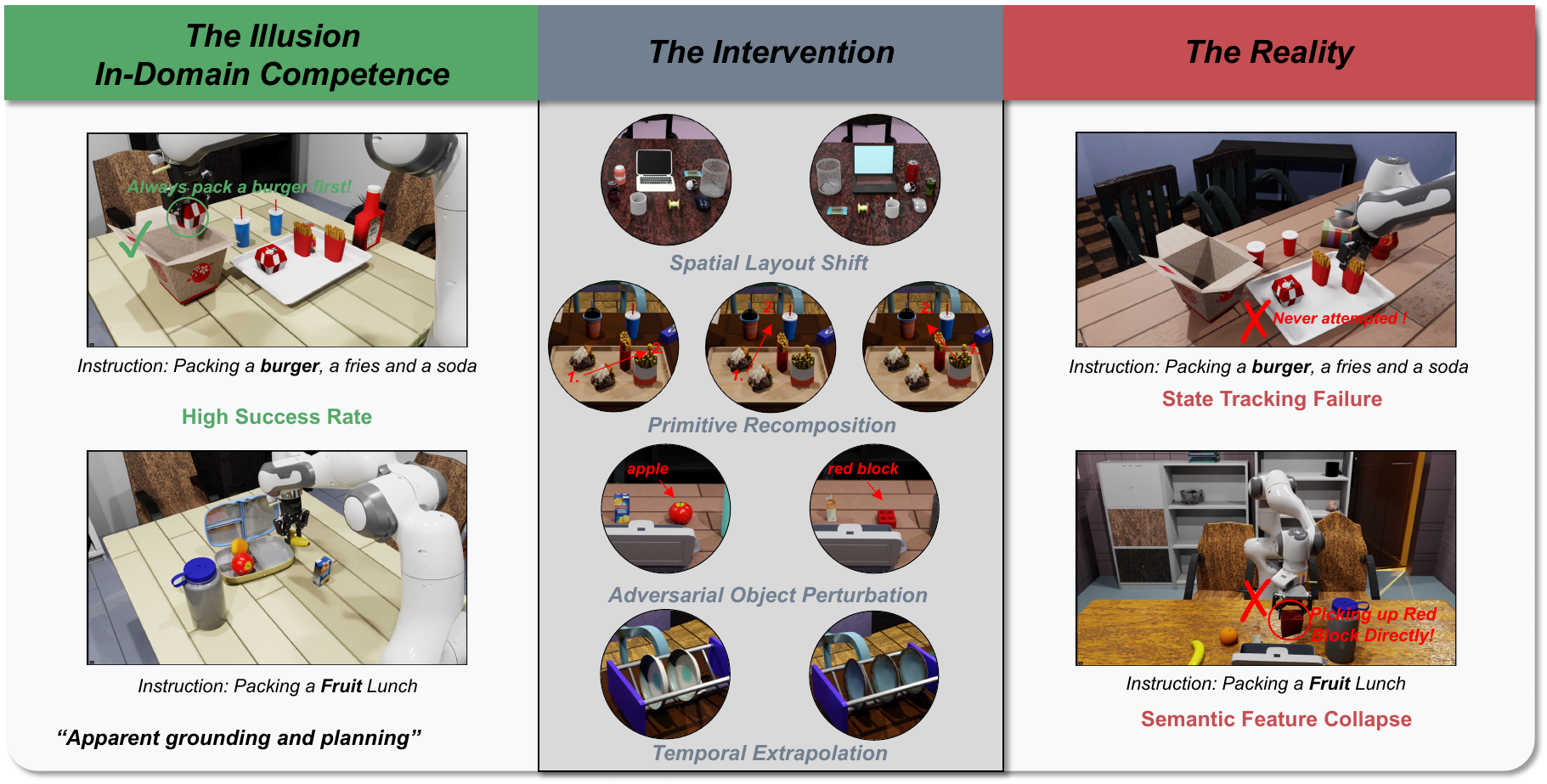}
 {\textbf{Unmasking the illusion of embodied reasoning.} \textbf{Left (Illusion):} State-of-the-art VLAs achieve high success rates on standard, visually static benchmarks, creating a misleading impression of robust semantic grounding and sequential planning. 
 \textbf{Middle (Intervention):} We introduce systematic cognitive stress tests via controlled interventions (e.g., spatial layout shifts and primitive recomposition) to challenge this apparent competence.
 \textbf{Right (Reality):} Performance collapses sharply, revealing failures in state tracking and semantic grounding, and exposing reliance on spurious correlations.}
  {fig:teaser}

\abstract{
Recent Vision-Language-Action (VLA) models report impressive success rates on standard robotic benchmarks, fueling optimism about general-purpose physical intelligence. 
However, recent evidence suggests a systematic misalignment between standard benchmark success and true embodied reasoning, raising the question of whether these high scores reflect genuine cognitive capability.
To address this gap, we introduce \textbf{BeTTER}, a diagnostic \textbf{Be}nchmark for \textbf{T}esting \textbf{T}rue \textbf{E}mbodied \textbf{R}easoning in robotic policies. BeTTER applies targeted causal interventions (e.g., spatial layout shifts, temporal extrapolation) while enforcing kinematic isolation to explicitly decouple high-level reasoning failures from low-level execution limits.
Through systematic evaluation, we reveal that state-of-the-art VLAs catastrophically fail in dynamic scenarios, exhibiting severe lexical-kinematic shortcuts, behavioral inertia, and semantic feature collapse.
Crucially, our mechanistic analysis traces these symptoms to fundamental architectural bottlenecks—such as capacity compression and myopic downsampling—which systematically degrade the model's foundational semantic representation.
We demonstrate that highly static evaluation protocols effectively mask this degradation by allowing optimization to overfit to sensorimotor priors.
Supported by real-world robotic validation, our findings confirm that this representational breakdown is not a simulation artifact, highlighting the critical need for future VLA paradigms to resolve the structural tension between high-frequency control and high-level reasoning.
}

\checkdata[Project Page]{\url{https://research.beingbeyond.com/better}}
\checkdata[Date]{Apr 20, 2026}

\begingroup
\footnotetext[2]{Correspondence to Zongqing Lu $<$lu@beingbeyond.com$>$.}
\endgroup

\begin{document}

\maketitle


\section{Introduction}

\label{sec:intro}

Over the past few years, the field has shifted its focus from visual understanding of human activities~\cite{carreira2017quo,zhao2017temporal,zheng2022few} to enabling robots to mimic and generalize human behaviors.
Recent advances in robotics~\cite{black2024pi0,nvidia2025gr00t, luo2026beingh05} have brought the field closer to achieving physical intelligence in embodied AI.
At the core of this vision lies general-purpose embodied reasoning, which means the ability to interpret complex instructions, reason about physical environments, and adapt skills across diverse tasks.
In particular, recent advances in robot policies, such as Vision-Language-Action (VLA) models~\cite{intelligence2025pi05, nvidia2025gr00t, luo2025being,luo2026beingh05}) have demonstrated strong performance on manipulation tasks requiring long-horizon execution and multi-step coordination.
In parallel, embodied QA-oriented VLMs, which can serve as the backbone of VLAs, also achieve impressive results on reasoning benchmarks~\cite{ji2025robobrain, hao2025mimo}.
Together, these advances suggest that reasoning-capable embodied agents may be closer than previously anticipated.

However, such progress can be misleading, as highlighted by recent studies~\cite{zhang2026vlm4vla}.
Existing evaluation protocols often reward behavioral competence without truly validating underlying reasoning.
Moreover, the supervision signals used in embodied reasoning benchmarks~\cite{sermanet2024robovqa, qiu2024egoplan, du2024embspatial, zhou2025roborefer} can improve benchmark performance without yielding corresponding gains in downstream action prediction.
As illustrated in Figure~\ref{fig:teaser}, this discrepancy gives rise to what we term the \emph{illusion of embodied reasoning}: policies appear competent in visually stable, in-domain settings, yet rely on shortcut correlations rather than genuine semantic grounding or causal reasoning.

Recent robustness-oriented benchmarks, such as LIBERO-Plus~\cite{fei2025liberoplus} and LIBERO-PRO~\cite{zhou2025liberopro}, take important steps toward evaluating policies under distribution shifts. 
However, their reliance on fixed simulation assets limits the ability to systematically vary task semantics, interaction structures, and underlying task logic.
This makes it difficult to disentangle reasoning failures from execution-level and perceptual confounders.
As VLAs become increasingly proficient at producing dense continuous actions, behavioral competence becomes easier to imitate, while true reasoning remains difficult to verify.


To address this gap, we introduce BeTTER, a diagnostic \textbf{Be}nchmark for \textbf{T}esting \textbf{T}rue \textbf{E}mbodied \textbf{R}easoning in robotic policies.
BeTTER constructs controlled task variations along multiple reasoning axes, including spatial layout shift, primitive recomposition, adversarial object perturbation, and temporal extrapolation.
Meanwhile, it logs privileged states to enable interpretable failure analysis. 

Across our diagnostic stress tests, state-of-the-art VLAs exhibit clear brittleness.
Despite near-saturated performance on standard benchmarks~\cite{liu2023libero}, they consistently fail under controlled reasoning interventions.
These failures manifest as recurring modes, including lexical-kinematic shortcuts, behavioral inertia, causal state tracking failures, and semantic feature collapse.

We further trace these limitations to a deeper architectural cause.
VLA models inherently face a trade-off between capacity compression and perceptual abstraction, leading to degraded semantic representations.
Consequently, policies are prone to shortcut learning during fine-tuning, overfitting to sensorimotor patterns instead of developing true reasoning capabilities.
This perspective resolves the apparent paradox between high benchmark performance and poor physical generalization, underscoring that genuine embodied reasoning requires more than mimicking dense continuous actions.
To validate that these findings are not artifacts of simulation, we conduct targeted stress tests on real-world robotic platforms, confirming that such representational failures persist in physical control settings.

Our contributions are summarized as follows:
\begin{itemize}
\item \textbf{An Extensible Diagnostic Benchmark:} 
We introduce BeTTER, a template-driven benchmark that generates controlled task variations to evaluate embodied reasoning while minimizing execution- and perception-level confounders.

\item \textbf{Diagnosing the Illusion of Embodied Reasoning:}
Through systematic evaluation, we uncover substantial reasoning deficits in state-of-the-art VLAs, revealing their reliance on visuospatial heuristics. 
We further trace these failures to fundamental representational bottlenecks arising during VLM-to-VLA adaptation, offering a mechanistic explanation for the resulting degradation in reasoning.
    
\item \textbf{Real-World Validation:}
We perform rigorous stress tests on a real-world robotic platform, demonstrating that the observed reasoning failures are not artifacts of simulation, but persist in physical control settings as a consequence of shortcut-driven optimization.
\end{itemize}

\section{Related Work}
\label{sec:related_work}

\noindent\textbf{Vision-Language-Action Models.}
The pursuit of general-purpose embodied agents has driven rapid progress in Vision-Language-Action (VLA) models, which aim to bridge high-level semantic reasoning with low-level motor control.
Early works~\cite{brohan2022rt, kim2024openvla, qu2025spatialvla}, such as RT-2~\cite{zitkovich2023rt} and OpenVLA~\cite{kim2024openvla}, formulate robotic control as a discrete next-token prediction problem~\cite{floridi2020gpt}, enabling emergent reasoning capabilities.
However, their scale introduces substantial computational overhead, limiting control frequency and hindering deployment in reactive, high-precision settings.
To address these limitations, recent approaches~\cite{black2024pi0, zheng2025x} adopt more efficient, high-frequency architectures, moving away from autoregressive action tokenization toward MLP heads~\cite{luo2025being} or diffusion-based control heads like flow matching~\cite{lipman2022flow}.
This shift enables faster motor control while attempting to retain the semantic priors learned from large-scale vision-language models~\cite{beyer2024paligemma, li2025eagle, zhang2025bpe}.
However, such compression can lead to the erosion of these priors.
To mitigate this, recent works~\cite{intelligence2025pi05,feng2025spatial} augment pre-training with dense spatial grounding and VQA datasets~\cite{yuan2024robopoint, ji2025robobrain}, aiming to preserve embodied reasoning capabilities beyond pure action imitation.

\noindent\textbf{Robot Manipulation Benchmarks.}
The rapid development of VLAs has also driven the evolution of evaluation protocols.
Early benchmarks, such as RLBench~\cite{james2020rlbench} and ManiSkill~\cite{mu2021maniskill}, established standards for low-level motor proficiency, while CALVIN \cite{mees2022calvin} introduced long-horizon, language-conditioned tasks for sequential instruction following.
To evaluate generalization across broader semantic and spatial contexts, LIBERO~\cite{liu2023libero} proposed a structured suite targeting knowledge transfer, and RoboCasa~\cite{nasiriany2024robocasa} introduced large-scale, everyday environments via generative simulation.
However, as models increasingly leverage large-scale real-world data~\cite{o2024open, bu2025agibot}, success rates on fixed hold-out sets become less informative.
High benchmark performance often fails to distinguish true reasoning from dataset interpolation, where models effectively memorizes visual-motor trajectories. 
This limitation highlights the need for diagnostic evaluation protocols that can systematically probe failure modes in these black-box policies.
Beyond scaling, recent efforts have focused on evaluating generalization under controlled perturbations.
Benchmarks such as SimplerEnv~\cite{li24simpler} and REALM \cite{sedlacek2025realm} target sim-to-real transfer, while others explicitly introduce structured variations.
For instance, GemBench~\cite{garcia2025towards} evaluates robustness to novel placements, object instances, and temporal extensibility, and VLABench~\cite{zhang2025vlabench} incorporates new object categories and implicit language instructions.
LIBERO-Plus~\cite{fei2025liberoplus} and LIBERO-Pro~\cite{zhou2025liberopro} further expose the fragility of current VLAs under visual perturbations.
To better characterize capability boundaries, VLA-Arena~\cite{zhang2025vlaarena} proposes a structured evaluation framework that disentangles task difficulty across task structure, language, and visual observation, enabling more fine-grained analysis of model behavior.

This work introduces BeTTER, shifting the diagnostic focus from perceptual robustness to embodied reasoning.
Rather than applying surface-level visual perturbations to static scenes, our extensible framework systematically manipulates the underlying task semantics and interaction logic, enabling more precise evaluation of reasoning capabilities in embodied agents.

\section{BeTTER: A Diagnostic Benchmark for True Embodied Reasoning}
\label{sec:benchmark}

Recent embodied benchmarks~\cite{fei2025liberoplus, zhou2025liberopro} introduce systematic perturbations, such as lighting, layout and texture, to improve evaluation robustness.
However, they remain constrained by fixed simulation assets, limiting the ability to generate meaningful out-of-distribution variations in task semantics, interaction structures, and underlying task logic. 

This limitation becomes increasingly pronounced as VLAs scale in capability, necessitating more diverse and targeted diagnostic scenarios for effective evaluation.
To address this gap, we introduce \textbf{BeTTER}, a \textbf{Be}nchmark for \textbf{T}esting \textbf{T}rue \textbf{E}mbodied \textbf{R}easoning in robotic policies through extensible diagnostic task variations.
Unlike traditional benchmarks that rely on static scenarios, \benchmark supports the dynamic incorporation of new 3D assets, programmatic task generation, and controlled diagnostic perturbations.
In addition, it logs privileged physical states during execution, enabling fine-grained behavioral analysis beyond simple success metrics. 
Together, these features provide a flexible testbed for probing the reasoning boundaries of embodied models especially VLAs and uncovering failure modes overlooked by existing benchmarks.
Specifically, \benchmark is instantiated with 10 base manipulation tasks spanning diverse physical constraints, from loose containment to precision insertion (see Appendix~\ref{sec:supp_evaluated_tasks} for a full taxonomy). These base tasks are systematically expanded into 60 variations through controlled perturbations along our diagnostic axes.

\subsection{Asset Curation and Modular Task Design}

\label{sec:asset_and_task}

To overcome the limitations of fixed simulation assets, the \benchmark pipeline enables dynamic construction of interactive environments.
Instead of relying on hard-coded scripts, tasks are defined through a modular, template-based design that supports scalable generation of diverse and semantically rich scenarios:

\textbf{1) Template-driven Task Generation}.
To efficiently construct evaluation tasks with coherent logical structure, semantic realism, and physical feasibility, we first define high-level templates that capture core interaction patterns (e.g., manipulation, placement, or sequential assembly).
We then use a powerful VLM (e.g., Gemini~\cite{comanici2025gemini}) to instantiate these templates into concrete task specifications.
As detailed in Appendix~\ref{sec:supp_prompt}, we design structured prompts that guides the VLM to generate diverse, physically plausible scenarios while explicitly estimating key physical properties (e.g., mass and size boundaries).

\textbf{2) Open-vocabulary Environment Construction}.
Rather than relying on a fixed set of predefined objects or manually curated assets, \benchmark employs an open-vocabulary retrieval module to source 3D assets.
Given object descriptions produced during task instantiation, assets are retrieved from large-scale repositories such as Objaverse~\cite{deitke2023objaverse}.
As shown in Table~\ref{tab:asset_retrieval}, a single VLM-generated description can map to multiple distinct 3D assets, enabling one-to-many retrieval that increases intra-task visual and geometric diversity.
Retrieved assets are automatically normalized (e.g., re-centering and collision bounding) to ensure compatibility with the simulation environment.
However, purely automated normalization may lack the precision required for complex, contact-rich interactions (e.g., fitting a plate into a dish rack).
To address this, we incorporate a human-in-the-loop verification step for targeted spatial refinements.

\textbf{3) Controlled Diagnostic Variations}.
Once a base scenario is instantiated, \benchmark generates diagnostic variations through controlled perturbations, enabling systematic evaluation of embodied reasoning capabilities.

\subsection{Data Collection and Privileged State Logging}
\label{sec:data_collection}
To populate instantiated tasks with high-quality execution data, \benchmark supports efficient trajectory collection alongside automatic logging of dense, privileged simulation states.

\textbf{1) Trajectory Amplification}. 
For each base task, we collect a small set of canonical human demonstrations via teleoperation.
Instead of recording demonstrations for every variation, we expand the dataset through procedural amplification (e.g., MimicGen \cite{mandlekar2023mimicgen}), which retargets demonstrations to novel configuration of object poses, spatial layouts, and object instances.
This enables scalable generation of diverse and physically valid manipulation trajectories with minimal human labor.

\textbf{2) Privileged State Logging and VQA Generation}. 
During execution, the simulator automatically records synchronized, privileged states at each timestep, including depth maps, precise 2D/3D bounding boxes, and object segmentation masks. 
Crucially, to mitigate spatial hallucination in purely VLM-based scene descriptions, we leverage these states as ground truth to construct geometrically accurate Visual Question Answering (VQA) annotations.
Representative examples covering spatial geometry, counting, and execution tracking are provided in Appendix~\ref{sec:supp_logging_vqa}.
To complement this structural accuracy with semantic diversity, we further augment the pipeline by generating open-ended scene descriptions and VQA pairs using VLMs.
Together, these multimodal annotations provide rich supervisory signals for diagnosing policy failures and enable the exploration of alternative learning paradigms.

\section{Unmasking the Illusion of Embodied Reasoning}
\label{sec:main_analysis}
Embodied reasoning, as claimed by recent VLAs~\cite{chen2025internvla, qu2025eo}, can obtain improvement through pre-training evidenced by high success rates on standard downstream manipulation tasks. 
However, it remains unclear for existing evaluation protocols to distinguish whether this apparent proficiency stems from true reasoning or the mere exploitation of shortcut behaviors memorized from training data.
To address this, we conduct a diagnostic study using our \benchmark benchmark, with the goal to identify the reasoning weakness overlooked in prior VLAs.

We evaluate three representative VLAs:
$\pi_{0.5}$ \cite{intelligence2025pi05}, GR00T-N1.6 \cite{nvidia2025gr00t}, and Being-H0.5 \cite{luo2026beingh05}. 
To systematically probe their reasoning limitations, we introduce four categories of targeted semantic and physical interventions at test time:
\textbf{1) Spatial Layout Shift}.
We rearrange object positions or alter their spatial relationships with nearby distractors, testing the model's ability to adapt to out-of-distribution configurations.
\textbf{2) Primitive Recomposition}.
We assess compositional generalization by recombining learned action primitives, e.g., evaluating $B \rightarrow C$ after training on $A \rightarrow B$ and $A \rightarrow C$. 
\textbf{3) Adversarial Object Perturbation}.
We replace targets or distractors with visually similar but semantically distinct objects, examining whether models rely on true semantic grounding rather than superficial visual cues.
\textbf{4) Temporal Extrapolation}.
We extend task horizons by chaining subgoals or introducing repeated actions, evaluating the model’s ability to maintain consistent state tracking and generalize across longer time horizons.
To preserve physical and logical validity, all perturbations are applied based on the specific semantics of each task.
Below, we present key findings that expose fundamental limitations of current VLAs in embodied reasoning.

\begin{figure}[h!]
    \centering
    \includegraphics[width=0.95\linewidth]{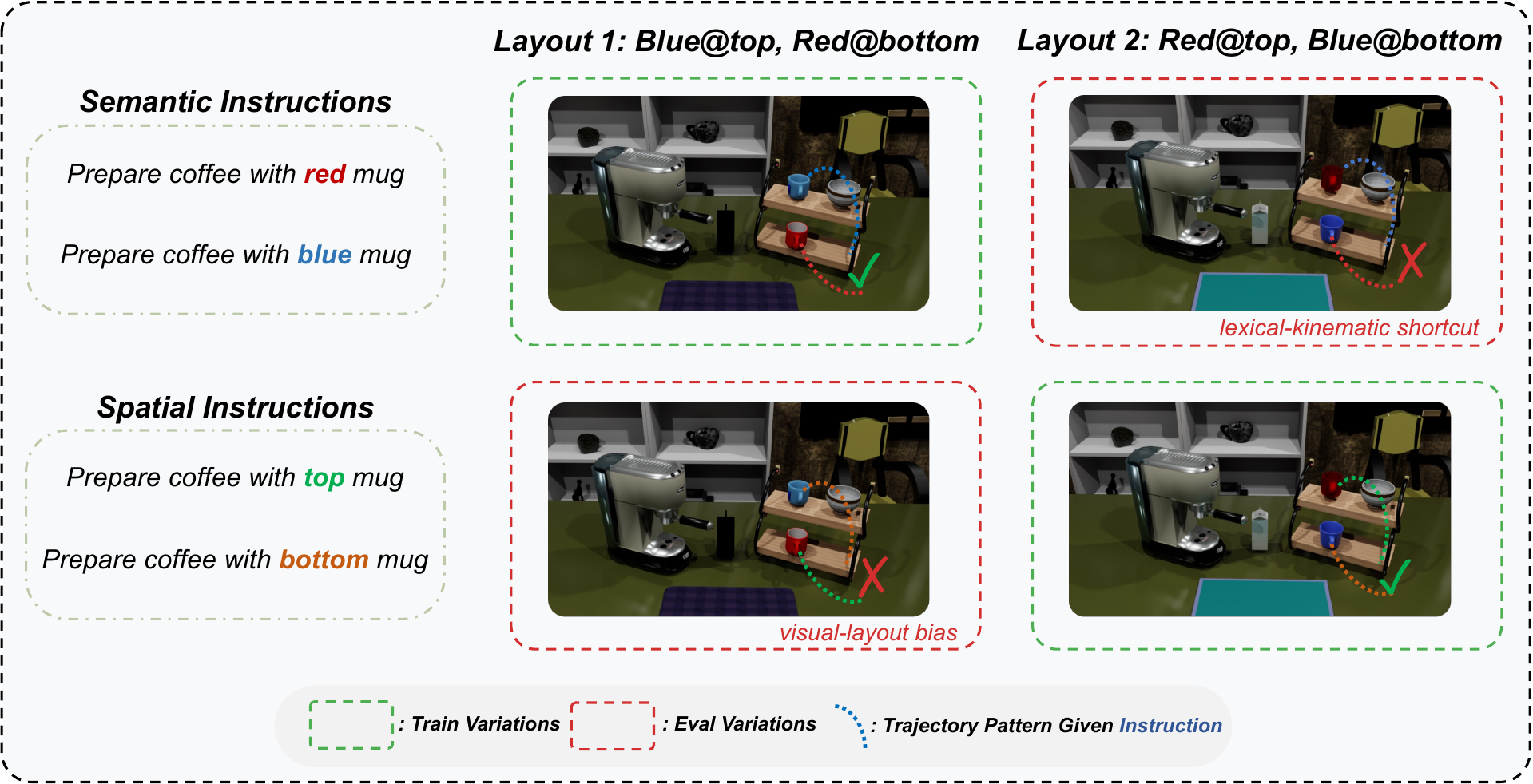}
    \caption{\textbf{Illustration of shortcut learning in instruction following.} 
    We evaluate models under counterfactual layouts by recombining spatial (\textit{top/bottom}) and semantic (\textit{red/blue}) instructions. 
    Dashed lines denote observed trajectory patterns. 
    Counterfactual cases (diagonal) expose two representative failure modes: 
    \textbf{Top-right} a lexical-kinematic shortcut, where the token ``red'' directly triggers a learned motion primitive ($move\_down$) without grounding in the visual scene. 
    \textbf{Bottom-left} a visual-layout bias, where the policy defaults to salient spatial positions (e.g., grasping the top object) while ignoring the semantic instruction.}
    \label{fig:instruction_following_wo_true_grounding}
\end{figure}

\begin{table}[h!]
\centering
\caption{\textbf{Quantitative results revealing apparent instruction understanding.} 
We report the success rate of grasping the correct target under specific layout-instruction pairings.
A $50\%$ random-guess baseline reflects the binary choice between two objects.
The pronounced performance polarization across variations reveals severe shortcut learning, indicating that models rely on spurious correlations rather than true visual–semantic grounding.
}
\label{tab:coffee_case}
\scalebox{.95}{
\setlength{\tabcolsep}{6pt}
\begin{tabular}{l | cc | cc }
    \toprule
    \multirow{2}{*}{\textbf{Model}} & \multicolumn{2}{c|}{\textbf{Spatial Inst. (Layout 1)}} & \multicolumn{2}{c}{\textbf{Semantic Inst. (Layout 2)}} \\
    & (``\textit{top}'') & (``\textit{bottom}'') & (``\textit{red}'') & (``\textit{blue}'') \\
    \midrule
    
    Random Guess & 50.0\% & 50.0\% & 50.0\% & 50.0\% \\
    \midrule
    $\pi_{0.5}$ & 65.0\% & 50.0\% & 70.0\% & 35.0\% \\
    GR00T-N1.6  & 100.0 \% & 100.0 \% & 5.0 \% & 5.0\% \\
    Being-H0.5  & 100.0 \% & 30.0\% & 95.0\% & 45.0\% \\
    \bottomrule
\end{tabular}}
\end{table}

\subsection{Finding 1: Apparent Instruction Understanding without True Grounding}
Recent VLAs demonstrate strong instruction-following on benchmarks~\cite{mees2022calvin}.
However, these benchmarks often rely on simple instructions and do not adequately control for cross-modal confounders.
As a result, it remains unclear whether models truly ground language in visual semantics or instead exploit spurious correlations between linguistic tokens and familiar visual cues.
To investigate this, we design a diagnostic task, \textit{Preparing Morning Coffee} (Figure~\ref{fig:instruction_following_wo_true_grounding}), where spatial relations (e.g., \textit{top/bottom}) and semantic attributes (e.g., \textit{red/blue}) are systematically recombined to form novel instruction–scene pairings.
All objects and locations are observed during training, ensuring that required motor primitives are already known.
This kinematic isolation guarantees that failures arise from breakdowns in instruction following rather than motion unfamiliarity.

As shown in Table~\ref{tab:coffee_case}, intention accuracies consistently deviate from random guessing ($50\%$), indicating that models are responsive to instructions.
However, the strong performance polarization across variations reveals significant shortcut learning.
Instead of aligning language with visual semantics, current VLAs tend to internalize spurious correlations present in the training data.

For example, GR00T-N1.6 achieves perfect accuracy on spatial instructions ($100\%$) but fails almost entirely on semantic ones ($5\%$), exhibiting a lexical-kinematic shortcut: the token ``red'' is directly mapped to a motor primitive (e.g., \textit{move\_down}) without visual verification. 
In contrast, Being-H0.5 shows a pronounced spatial location bias, performing well on ``top'' ($100\%$) and ``red'' ($95\%$ when located at the top) but failing on ``bottom'' and ``blue'', often defaulting to grasping the upper object regardless of the instruction.

Interestingly, $\pi_{0.5}$ degrades to near-random performance, frequently exhibiting early halting behavior. 
We hypothesize that this arises from conflicting multimodal conditioning.
During training, ``red'' is associated with a downward motion, but when the red object is moved to the top, the ``move down'' instruction clashes with the visual cues to move up. 
In continuous generative policies, such opposing signals may cancel out, resulting in near-zero velocity prediction and causing the robot to stall until stochastic noise breaks the deadlock.

\begin{figure}[h]
\centering
\includegraphics[width=.955\linewidth]{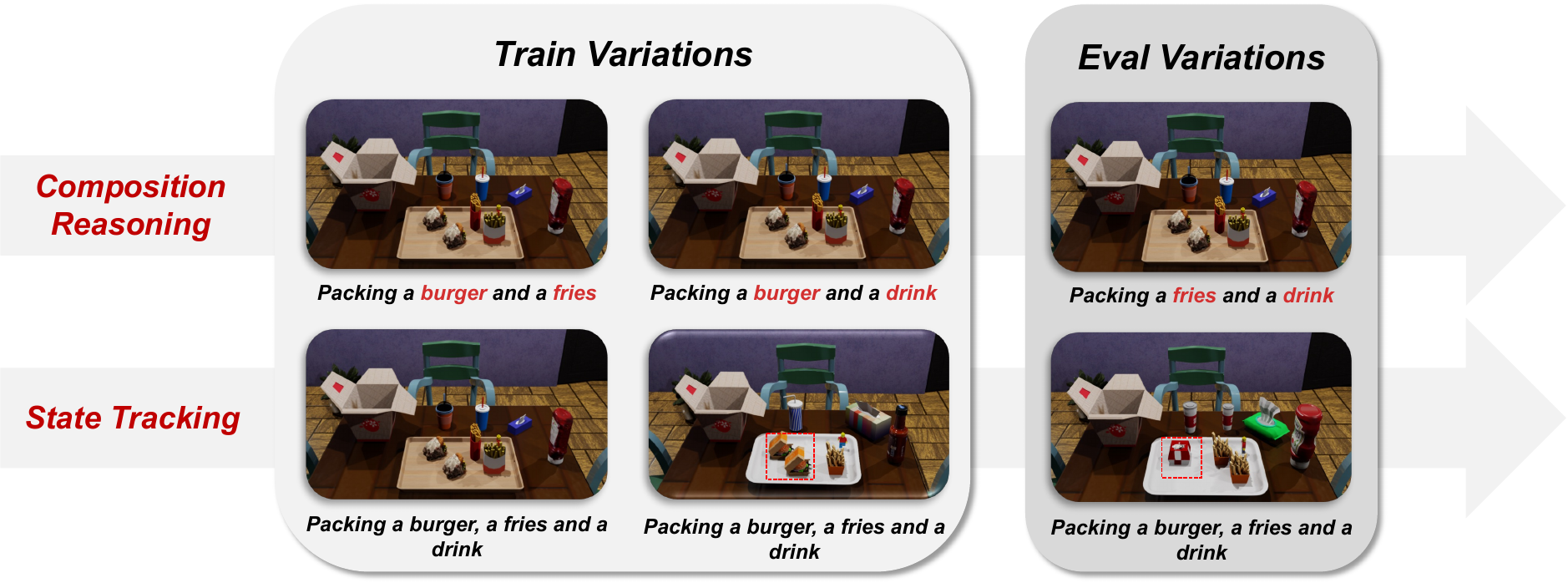}
\caption{
\textbf{Illustration of reasoning breakdowns in sequential planning.} 
We use the \textit{Packing Fast Food Order} task to expose two distinct failure modes.
\textbf{Top (Subgoal recomposition):} 
Models trained on fixed sequences ($A\rightarrow B$ and $A\rightarrow C$) fail to generalize to novel compositions ($B\rightarrow C$), instead exhibiting behavioral inertia by defaulting to the most frequent starting primitive ($A$).
\textbf{Bottom (Causal confusion):} 
Models memorize the visual cue ``one burger remaining'' as a trigger for ``pack fries''.
When the episode begins with only one burger, they misinterpret the initial state as mid-progress, hallucinate task advancement, and incorrectly skip the first step.
}
\label{fig:fast_food_variation}
\end{figure}

\subsection{Finding 2: Failure of Subgoal Composition Generalization}
To evaluate whether models can flexibly decompose and recombine learned behaviors, we design the \textit{Packing Fast Food Order} task (Figure~\ref{fig:fast_food_variation}, Top).
This task requires zero-shot recomposition of subgoals under novel temporal orders.
During training, models observe sequences such as $A \rightarrow B$ (e.g., pack burger, then fries) and $A \rightarrow C$ (e.g., pack burger then drink). 
At test time, they are required to execute a novel composition $B \rightarrow C$ (e.g., pack fries, then drink), which cannot be solved by replaying any previously seen trajectory.
As shown in Table~\ref{tab:composition_case}, performance drops sharply on these unseen compositions,revealing a systematic failure we term \textit{behavioral inertia}.
Because all training sequences begin with packing the burger ($A$), models acquire a strong unconditional bias toward this initial primitive.
When presented with the novel instruction $B \to C$, the policy collapses to unconditional behavioral cloning, mapping the initial visual state (a full tray of food) to the most common action (grasping the burger), while ignoring the compositional structure of the instruction.
This behavior mirrors the visual-layout bias observed in Finding 1, suggesting that the illusion of embodied reasoning extends beyond spatial grounding to temporal planning and subgoal composition.

\begin{table}[h]
    \centering
    \caption{
\textbf{Quantitative results on subgoal composition generalization.} 
We evaluate models’ ability to recompose learned action primitives under novel task structures.
While models memorize the seen training sequences ($A \rightarrow B$ and $A \rightarrow C$), performance drops sharply on the unseen composition ($B \rightarrow C$). 
$\Delta \text{SR}$ denotes the absolute decrease in success rate, highlighting the reliance on memorized trajectory patterns rather than compositional reasoning over instructions.
}
    \label{tab:composition_case}
    \scalebox{.95}{
    \setlength{\tabcolsep}{5pt}
    \begin{tabular}{l | cc | c | c }
        \toprule
        \multirow{2}{*}{\textbf{Model}} & \multicolumn{2}{c|}{\textbf{Seen Subgoals}} & \textbf{Novel Composition} & \multirow{2}{*}{\textbf{$\Delta$ SR}} \\
        & ($A \rightarrow B$) &  ($A \rightarrow C$) & \textit{Test Var} ($B \rightarrow C$) & \\
        \midrule
        $\pi_{0.5}$ & 60.0 & 45.0 & 5.0 & \textcolor{red}{-47.5} \\
        GR00T-N1.6  & 75.0 & 40.0 & 15.0 & \textcolor{red}{-42.5} \\
        Being-H0.5  & 65.0 & 40.0 & 0.0 & \textcolor{red}{-52.5}\\
        \bottomrule
    \end{tabular}}
\end{table}

\subsection{Finding 3: Failure of Causal State Tracking.}
A robust reasoning agent must maintain an accurate internal representation of task progress.
However, imitation-based policies often rely on superficial visual cues as proxies for true causal state tracking.
To expose this limitation, we introduce targeted variations in the \textit{Packing Fast Food Order} task (Figure~\ref{fig:fast_food_variation}, Bottom).
During training, all episodes begin with two burgers on the table, and the operator consistently packs one first.
Consequently, the visual state ``one burger remaining'' becomes mistakenly tied with the next subgoal, ``pack the fries''.
At evaluation time, we modify the initial condition to contain only one burger, while keeping the instruction unchanged. 
Instead of executing the correct first step (packing the burger), models skip it and directly attempt to pack the fries. 
This behavior reveals a form of causal confusion, where models fail to track which subgoals have actually been completed.
Rather than maintaining an explicit notion of task progress, they rely on superficial visual patterns (e.g., remaining objects) as shortcuts for inferring state.
More broadly, this highlights a critical diagnostic insight: when training data exhibits consistent but non-causal state-action correlations, VLAs internalize them as rigid decision rules.
As a result, policies may hallucinate task progress and execute incorrect actions, despite having access to the correct motor primitives and clear semantic instructions.

\begin{figure}[t]
    \centering
    
    \begin{subfigure}[b]{0.32\linewidth}
        \centering
        \includegraphics[width=\linewidth]{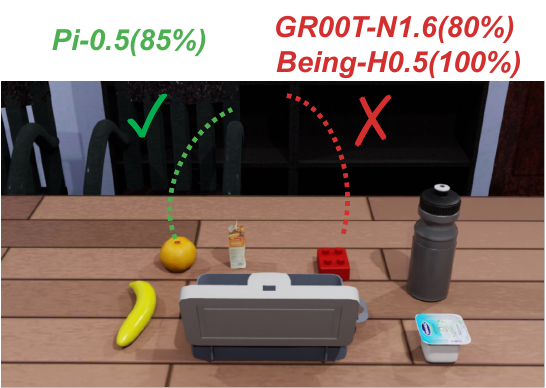}
        \caption{Instance-Level Distractor}
        \label{fig:sub_instance}
    \end{subfigure}
    \hfill
    \begin{subfigure}[b]{0.32\linewidth}
        \centering
        \includegraphics[width=\linewidth]{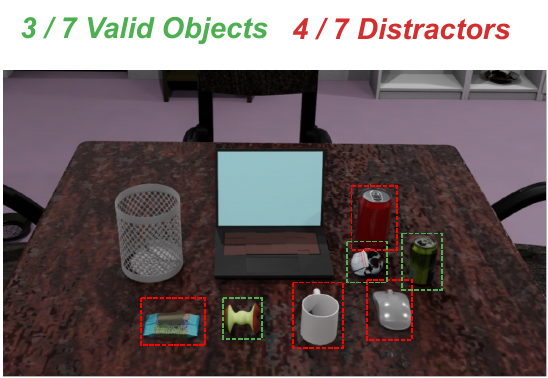}
        \caption{Scene-Level Clutter}
        \label{fig:sub_scene}
    \end{subfigure}
    \hfill
    \begin{subfigure}[b]{0.32\linewidth}
        \centering
        \includegraphics[width=\linewidth]{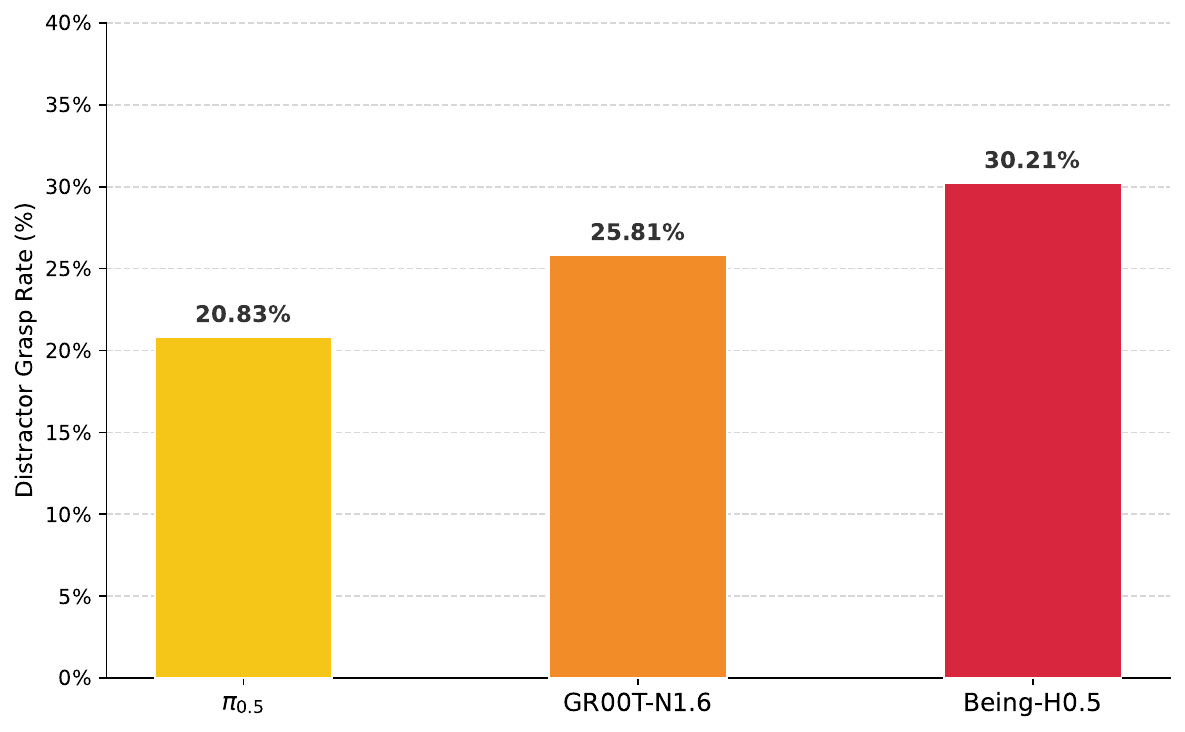}
        \caption{Distractor Grasp Rate}
        \label{fig:sub_dgr}
    \end{subfigure}

\caption{
\textbf{Evaluation of fine-grained semantic grounding under adversarial visual distractors.}
\textbf{(a) Instance-Level:} Adversarial substitutions (e.g., replacing a red apple with a red block) induce severe semantic feature collapse, leading models to act on superficial attributes such as color, $\pi_{0.5}$ shows partial robustness.
\textbf{(b) Scene-Level:} 
In cluttered environments without reliable spatial priors, policies fail to ground target semantics (e.g., trash) and instead interact with functionally irrelevant distractors (e.g., a computer mouse).
\textbf{(c) Metric (DGR):}
The Distractor Grasp Rate (DGR) quantifies this failure when spatial familiarity is removed, policies abandon semantic discrimination and regress to a ``blind grasping'' heuristic, selecting any salient object.
}
    \label{fig:finding_semantic_boundary}
\end{figure}

\subsection{Finding 4: Failure of Fine-Grained Semantic Grounding.}
A truly generalizable policy must possess high-resolution semantic understanding to distinguish between objects that share superficial visual attributes (e.g., color or shape) but differ in identity or affordance. 
However, current embodied policies often exhibit semantic feature collapse, relying on low-level visual salience rather than robust object grounding.

We first examine this at the instance level.
In the \textit{Packing a Fruit Lunch} task, the target ``red apple'' is replaced with a visually similar ``red block'' at the same location (Figure~\ref{fig:sub_instance}).
Policies frequently execute invalid grasps on the distractor, reflecting a combination of spatial memorization and semantic collapse.
Notably, $\pi_{0.5}$ shows partial robustness, correctly rejecting the distractor in 85\% of trials, suggesting a baseline level of semantic discrimination in simple, uncluttered settings.
However, this robustness does not extend to more complex environments.
In the \textit{Tidying Office Desk} task, policies must distinguish valid targets (e.g., crumpled napkins or cans) from functional distractors (e.g., computer mice or intact mugs) (Figure~\ref{fig:sub_scene}). 

To isolate semantic grounding from spatial priors, we evaluate only novel layouts.
In these settings, the illusion of semantic understanding breaks down.
We quantify this degradation using the \textbf{Distractor Grasp Rate (DGR)}, defined as the proportion of grasp attempts directed at invalid objects (excluding null or aborted actions).
As shown in Figure~\ref{fig:sub_dgr}, all models exhibit substantial DGRs under unseen layouts.
While such error rates may appear moderate in classification tasks, they are catastrophic in physical deployment --- for example, discarding a functional computer mouse instead of a crumpled napkin.
These results demonstrate that, under clutter and without spatial priors, current policies fail to maintain fine-grained semantic grounding and instead revert to a ``blind grasping'' heuristic, interacting with any salient object within reach regardless of task intent.

\section{Root Causes of Embodied Reasoning Degradation}
\label{sec:degradation}

The diagnostic findings above reveal a pervasive ``illusion of embodied reasoning'': under targeted causal interventions, state-of-the-art VLAs fail to exhibit genuine semantic grounding and sequential planning.
Instead, they systematically regress to fragile kinematic memorization and superficial visual heuristics.
This exposes a fundamental paradox: although these models are initialized from powerful vision-language models with strong reasoning capabilities, such high-level cognition consistently degrades during continuous physical control.
Why does this gap emerge?
To answer this question, we must look beyond observed behaviors and examine the representational bottlenecks inherent in the VLA deployment lifecycle.

\subsection{What Does Embodied Reasoning Require?}
To generalize in open-world environments, a robotic policy must go beyond robust low-level sensorimotor control and acquire a core set of cognitive capabilities: understanding complex spatial topologies, grounding fine-grained object semantics, and inferring causal states for sequential planning. 
We formalize these capabilities as a unified visual–semantic and causal–relational representation, denoted as $z$.
During execution, this high-level representation $z$ must operate in conjunction with the policy’s sensorimotor and kinematic prior $z_a$, which governs reactive, high-frequency continuous control.
Effective embodied reasoning thus depends on the coherent integration of $z$ (cognition) and $z_a$ (action).

Importantly, the three mentioned cognitive dimensions of $z$ align closely with the evaluation axes of static embodied VLM benchmarks, including EmbSpatial~\cite{du2024embspatial}, RefSpatial~\cite{zhou2025roborefer}, and EgoPlan-Bench2~\cite{qiu2024egoplan}. 
This correspondence suggests that static VQA-style evaluation is not orthogonal to continuous control.
Rather, it serves as a valuable diagnostic proxy for assessing whether the high-level representation $z$ remains intact, before being degraded by representational bottlenecks during VLA deployment.

\subsection{Where Does the Reasoning Go?}
\label{sec:vlm_to_vla_degradation}
To empirically investigate the degradation of representation $z$, we trace the architectural and perceptual compromises that a foundational vision-language Model undergoes when adapted for real-time VLA deployment.
To isolate these factors from confounding effects of continuous control, we perform static evaluations on the aforementioned reasoning benchmarks.
As shown in Table~\ref{tab:static_bottleneck}, we first establish performance upper bounds using state-of-the-art VLMs.
We then progressively ablate a strong base model (InternVL-3.5) to simulate the VLA deployment, enabling a controlled analysis of how representational capacity degrades across stages, with the key findings as follows:

\begin{table*}[t]
    \centering
\caption{
\textbf{Step-by-step degradation of reasoning in the VLM-to-VLA transition.} \textbf{Top:} Reference upper bounds from generalist and embodied VLMs.
\textbf{Bottom:} Ablations tracing the architectural compromises required for VLA deployment based on InternVL-3.5.
The results reveal a cascading collapse of the reasoning representation $z$:
1) \textbf{Capacity scaling} (8B $\rightarrow$ 2B) significantly weakens foundational reasoning ability; 
2) \textbf{VLA co-training} partially recovers spatial grounding but saturates the limited representation space, failing to improve sequential planning; 
3) \textbf{Myopic perceptual constraints} (enforcing a single $224 \times 224$ input for real-time control) introduce a hard bottleneck, erasing fine-grained semantics necessary for robust embodied reasoning.
}
    \label{tab:static_bottleneck}
    \scalebox{.90}{
    \setlength{\tabcolsep}{2.5pt}
    \begin{tabular}{l | c | ccc | ccccc }
        \toprule
        \multirow{2}{*}{\textbf{Model / Configuration}} & 
        \multirow{2}{*}{\textbf{EmbSp.}} & 
        \multicolumn{3}{c|}{\textbf{RefSpatial}} & 
        \multicolumn{5}{c}{\textbf{EgoPlan-Bench2}} \\
        
        \cmidrule(lr){3-5} \cmidrule(lr){6-10}
        & & Loc. & Plc. & All & Daily & Hob. & Rec. & Work & All \\
        \midrule
        
        \multicolumn{10}{l}{\cellcolor{gray!10}\textit{Reference Upper Bounds: Generalist \& Embodied VLMs}} \\
        Gemini-2.5-Pro-preview-05-06 & 78.74 & 44.58 & 31.73 & 38.16 & 44.19 & 43.05 & 46.45 & 39.60 & 42.85 \\
        GPT-4o-2024-11-20 & 71.92 & 8.00 & 9.55 & 8.78 & 47.38 & 40.00 & 44.81 & 35.64 & 41.79 \\
        Claude-Sonnet-4-2025-05-14 & 64.26 & 5.00 & 10.37 & 7.69 & 43.51 & 41.02 & 42.62 & 38.87 & 41.26 \\
        Mimo-Embodied \cite{hao2025mimo} (7B)  & 76.24 & - & - & 48.00 & - & - & - & - & 43.00 \\
        RoboBrain-2.0 \cite{team2025robobrain} (7B) & 76.32 & 36.00 & 29.00 & 32.50 & 39.41 & 32.20 & 33.88 & 26.98 & 33.23\\
        RoboBrain-2.0 \cite{team2025robobrain} (32B) & 78.57 & 54.00 & 54.00 & 54.00 & 64.01 & 53.22 & 57.92 & 52.48 & 57.23 \\
        
        \midrule
        \multicolumn{10}{l}{\cellcolor{gray!10}\textit{Ablating VLA Constraints (Base: InternVL-3.5 \cite{wang2025internvl3.5})}} \\
        \textbf{1. Scaling down Capacity} & & & & & & & & & \\
        \hspace{3mm} \textbf{8B} Base    & 74.96 & 48.00 & 7.00 & 27.50 & 47.15 & 39.66 & 44.26 & 33.17 & 40.80 \\
        \hspace{3mm} \textbf{4B} Base     & 72.08 & 38.00 & 8.00 & 23.00 & 38.50 & 40.55 & 38.25 & 31.67 & 36.83 \\
        \hspace{3mm} \textbf{2B} Base     & 60.19 & 10.00 & 2.00 & 6.00  & 36.90 & 33.56 & 36.07 & 28.22 & 33.38 \\
        \midrule
        \textbf{2. The Pre-training Shift} & & & & & & & & & \\
        \hspace{3mm} \textbf{2B + VLM + VLA Data} (Multi-patch) & 57.47 & 41.00 & 18.00 & 29.50 & 37.36 & 29.83 & 36.07 & 25.74 & 31.95 \\
        \midrule
        \textbf{3. The "Myopic" Perceptual Constraint} & & & & & & & & & \\
        \hspace{3mm} \textbf{2B + VLM + VLA Data (Single 224px)} & 52.42 & 9.00 & 6.00 & 7.50 & 37.36 & 29.83 & 36.07 & 25.74 & 31.95 \\
        \bottomrule
    \end{tabular}}
\end{table*}

\noindent\textbf{Degradation from Capacity Compression.}
A primary compromise in VLA deployment is the need to compress model capacity to meet onboard computational limits. 
As shown in Table~\ref{tab:static_bottleneck}, reducing the base model from 8B to 2B parameters leads to substantial performance degration.
The 2B model exhibits pronounced drops in both high-level sequential reasoning (EgoPlan 40.80 $\rightarrow$ 33.38) and semantic grounding (RefSpatial 27.50 $\rightarrow$ 6.00). This compression imposes a strict upper bound on the model’s representational bandwidth, limiting its ability to sustain rich visual–semantic and causal reasoning.

\noindent\textbf{Co-Training Asymmetry between Perception and Planning.}
To adapt compressed VLMs for embodied control, VLAs are co-trained on mixed datasets combining embodied reasoning (VLM data) and robotic interaction (VLA data).
However, this process introduces a co-training asymmetry. 
While large-scale spatial grounding data can partially recover perceptual localization (RefSpatial 6.00 $\rightarrow$ 29.5), this gain comes at the expense of high-level reasoning capacity.
Sequential planning performance not only fails to improve but slightly regresses relative to the 2B base model (EgoPlan 33.38 $\rightarrow$ 31.95). 
This suggests that the limited representational space of the compressed model becomes dominated by low-level sensorimotor features ($z_a$), leaving insufficient capacity to preserve causal–relational reasoning ($z$).
As a result, the policy degenerates into a reactive perceiver rather than a planner.

\noindent\textbf{Degradation from ``Myopic'' Perceptual Constraint.}
Even with co-training, a critical mismatch remains between VLM evaluation and VLA deployment.
Modern VLMs process high-resolution inputs as multiple image patches (e.g., 1–6 patches of $448 \times 448$), preserving fine-grained spatial detail through thousands of visual tokens.
In contrast, real-time VLAs must downsample observations to a single low-resolution input (e.g., $224 \times 224$) to meet control frequencies.
We simulate this constraint by enforcing a ``myopic'' perceptual setting on the pre-trained model.
As shown in the final row of Table~\ref{tab:static_bottleneck}, this structural bottleneck eliminates previously recovered spatial grounding performance (RefSpatial 29.50 $\rightarrow$ 7.50). 
This result directly links the semantic collapse and ``blind grasping'' observed in Section \ref{sec:main_analysis} to the loss of perceptual fidelity under real-time constraints, particularly in cluttered environments.

\subsection{Why Do Degraded Models Still Pass Benchmarks?}
\label{sec:sterile_bubble_and_validation}

\begin{table}[t]
\centering
\caption{
\textbf{Resolving the benchmark paradox through out-of-distribution (OOD) evaluation.} 
\textbf{Top:} State-of-the-art reference models. 
\textbf{Bottom:} Ablations of our pre-training recipes. 
On the highly static LIBERO benchmark, all configurations, including the ``No Pre-training'' baseline, saturate near $97\%$ success, indicating that in-distribution tasks can be solved primarily through kinematic representation $z_a$. 
In contrast, under CALVIN's OOD split (ABC$\to$D), environmental shifts invalidate these kinematic heuristics.
In this regime, incorporating diverse VLM data (``VLA + VLM'') helps preserve the underlying semantic representation $z$, leading to a substantial improvement in sequential task completion compared to pure continuous-control pre-training (``VLA Only'').
}
\label{tab:combined_results}
    
    \scalebox{.95}{
    \setlength{\tabcolsep}{3.5pt} 
    \begin{tabular}{l ccccc ccccc c}
        \toprule
        \multirow{3}{*}{\textbf{Configuration}} & 
        \multicolumn{5}{c}{\textbf{LIBERO}} & 
        \multicolumn{6}{c}{\textbf{CALVIN (ABC $\to$ D)}} \\
        
        \cmidrule(lr){2-6} \cmidrule(lr){7-12}
        
        & \multirow{2}{*}{\textbf{Spatial}} & \multirow{2}{*}{\textbf{Object}} & \multirow{2}{*}{\textbf{Goal}} & \multirow{2}{*}{\textbf{Long}} & \multirow{2}{*}{\textbf{Avg}} & \multicolumn{5}{c}{\textbf{Task completed in a row}} & \multirow{2}{*}{\textbf{Avg. Len. $\uparrow$}} \\
        
        \cmidrule(lr){7-11}
        
        & & & & & & \textbf{1} & \textbf{2} & \textbf{3} & \textbf{4} & \textbf{5} & \\
        \midrule
        
        \rowcolor[gray]{0.95} \multicolumn{12}{l}{\textit{Established Baselines (Reference Models)}} \\
        GR00T-N1.6 \cite{nvidia2025gr00t} & 97.65 & 98.45 & 97.5 & 94.35 & 96.99 & 96.3 & 91.4 & 85.3 & 79.7 & 71.7 & 4.244 \\ 
        $\pi_{0.5}$ \cite{intelligence2025pi05} & 98.8 & 98.2 & 98.0 & 92.4 & 96.85 & 93.5 & 86.8 & 81.4 & 77.5 & 73.4 & 4.126 \\
        Being-H0.5 \cite{luo2026beingh05} & 99.2 & 99.6 & 99.4 & 97.4 & 98.9 & 93.4 & 87.0 & 82.0 & 77.7 & 73.7 & 4.138 \\
        
        \midrule
        \rowcolor[gray]{0.95} \multicolumn{11}{l}{\textit{Pre-training Recipe Ablations (Ours)}} \\
        No Pre-training & 97.8 & 98.8 & 96.4 & 96.6 & 97.4 & 89.9 & 80.1 & 71.8 & 66.4 & 58.7 & 3.669 \\ 
        VLA (OXE Only) & 97.2 & 98.4 & 97.2 & 92.2 & 96.25 & 93.1 & 84.6 & 76.2 & 69.4 & 62.7 & 3.860 \\ 
        VLA + VLM & 96.6 & 97.6 & 98.2 & 96.4 & 97.2 & 93.5 & 85.9 & 81.0 & 77.1 & 71.1 & 4.086 \\ 
        \bottomrule
    \end{tabular}}
\end{table}

A key paradox emerges: if state-of-the-art VLAs suffer from significant reasoning degradation, why do they still achieve near-perfect performance on standard robotic benchmarks?
We argue that this illusion lies in the interaction between benchmark environment distributions and the optimization dynamics of downstream fine-tuning.
To validate this hypothesis, we analyze the impact of different pre-training data recipes (detailed in Appendix~\ref{sec:supp_libero_calvin}) across both in-distribution (LIBERO~\cite{liu2023libero}) and out-of-distribution (CALVIN ABC$\to$D~\cite{mees2022calvin}) settings.

\noindent\textbf{In-Distribution Saturation Masks Degradation.}
As shown in Table~\ref{tab:combined_results}, on the highly static LIBERO benchmark, both the ``VLA Only'' and ``VLA + VLM'' configurations achieve saturated success rates ($\sim$97\%).
In visually consistent, fixed-camera environments, tasks exhibit limited spatial and semantic variation. 
As a result, benchmark success cannot distinguish true semantic reasoning from trajectory memorization.

\noindent\textbf{Out-of-Distribution Evaluation Requires Semantic Grounding.}
The gap becomes evident under CALVIN's out-of-distribution split.
Unseen environmental variations invalidate memorized spatial heuristics, requiring robust semantic representation $z$ that LIBERO does not test.
Under these conditions, the ``VLA + VLM'' configuration significantly outperforms the ``VLA Only'' baseline (Avg. Len.: 3.860 $\rightarrow$ 4.086), highlighting the importance of preserved semantic grounding.

\noindent\textbf{Mechanism: Optimization Dynamics During Fine-Tuning.}
These contrasting results can be explained by the optimization dynamics of the fine-tuning stage.
When training on pure action demonstrations, the learning process naturally favors representational pathways that minimize immediate action-prediction loss.

If the semantic representation $z$ has already degraded due to architectural bottlenecks (Section \ref{sec:vlm_to_vla_degradation}), gradients will disproportionately update the remaining sensorimotor and kinematic representation $z_a$. 
This induces a severe form of \textit{shortcut learning}: instead of grounding actions in visual semantics, the policy increasingly relies on proprioceptive signals, effectively bypassing visual reasoning.
Such spatial overfitting is highly rewarded in static environments like LIBERO, but fails catastrophically under distribution shifts, as in CALVIN.

In contrast, the ``VLA + VLM'' recipe incorporates diverse multimodal VLM data during pre-training, which regularizes and partially preserves the representation $z$. 
As a result, during fine-tuning, the policy can anchor action generation to stable visual semantics, enabling it to adapt to novel environments through semantic reasoning rather than memorized kinematic patterns.

\subsection{Discussion of VLA Design}
\label{sec:takeaway_vla_paradox}

The preceding diagnostics and empirical analyses reveal a fundamental tension in designing generalist robotic agents: balancing high-level reasoning capacity with the latency constraints of high-frequency control.
Our results show that robust visual–semantic reasoning and causal planning (i.e., a strong $z$) benefit from expressive backbone capacity and high-resolution, multi-patch visual inputs.
In contrast, reactive continuous control demands low-latency inference at high control frequencies.
When these competing requirements are unified within a single end-to-end Vision–Language–Action model, optimization is typically dominated by execution efficiency.
This leads to degraded perception, representational collapse, and the shortcut learning behaviors identified throughout this work.
These findings suggest that current VLA paradigms implicitly trade away semantic reasoning for control efficiency. 
Future approaches to physical intelligence must instead introduce structural mechanisms to decouple or preserve high-level representations, ensuring that the semantic and causal reasoning capacity is maintained rather than sacrificed for real-time motor execution.

\section{Real-World Experiments}
\label{sec:real_world_experiment}

\begin{figure*}[t]
    \centering
    \begin{subfigure}[b]{0.24\textwidth}
        \centering
        \includegraphics[width=\textwidth]{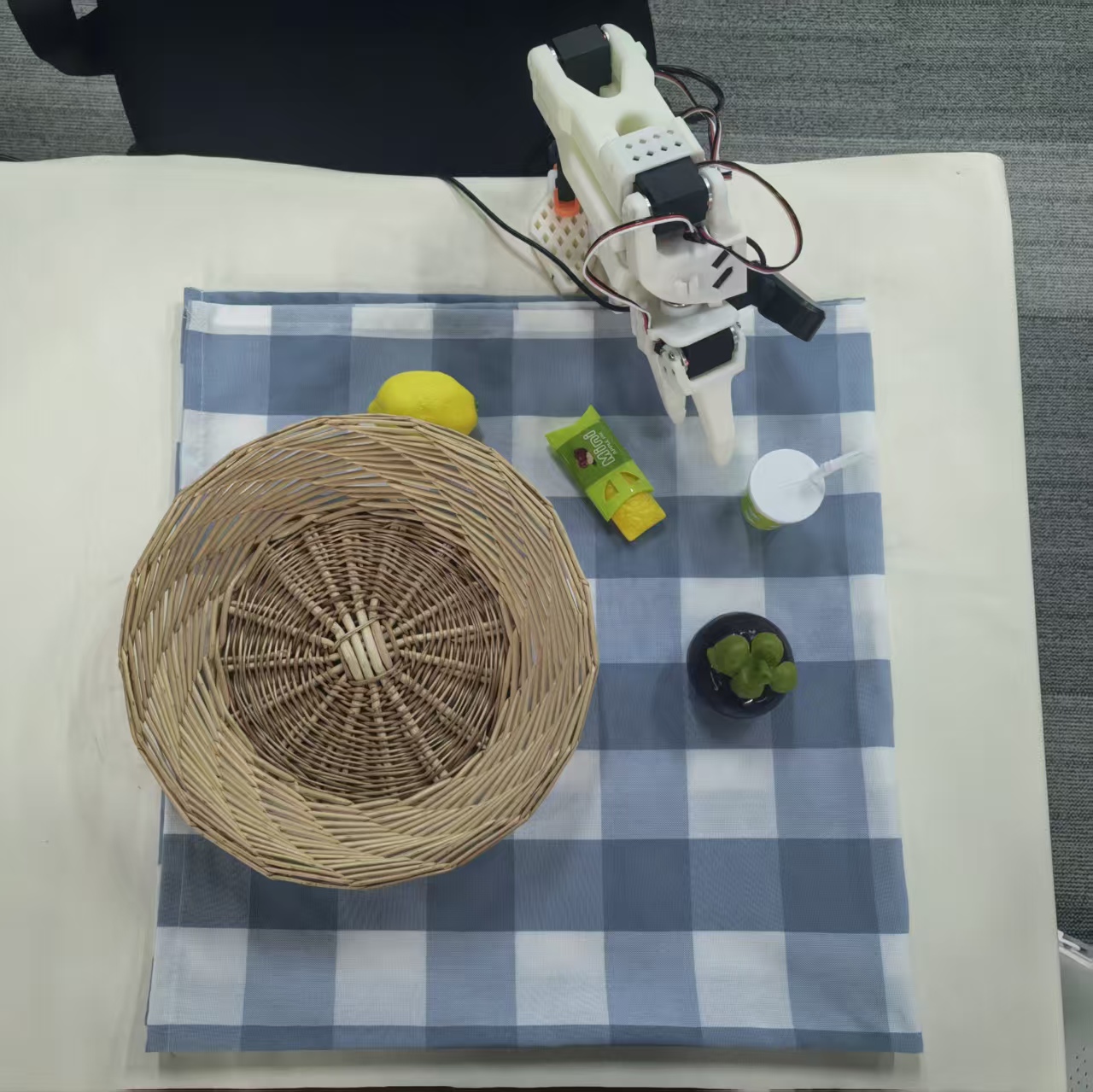}
        \caption{\begin{tabular}[t]{@{}c@{}} 
            \textbf{Train: Canonical} \\ 
            Short-horizon \\ 
            Layout 1 \\ 
            Blue tablecloth 
        \end{tabular}}
        \label{fig:train_t1_l1_b}
    \end{subfigure}
    \hfill
    \begin{subfigure}[b]{0.24\textwidth}
        \centering
        \includegraphics[width=\textwidth]{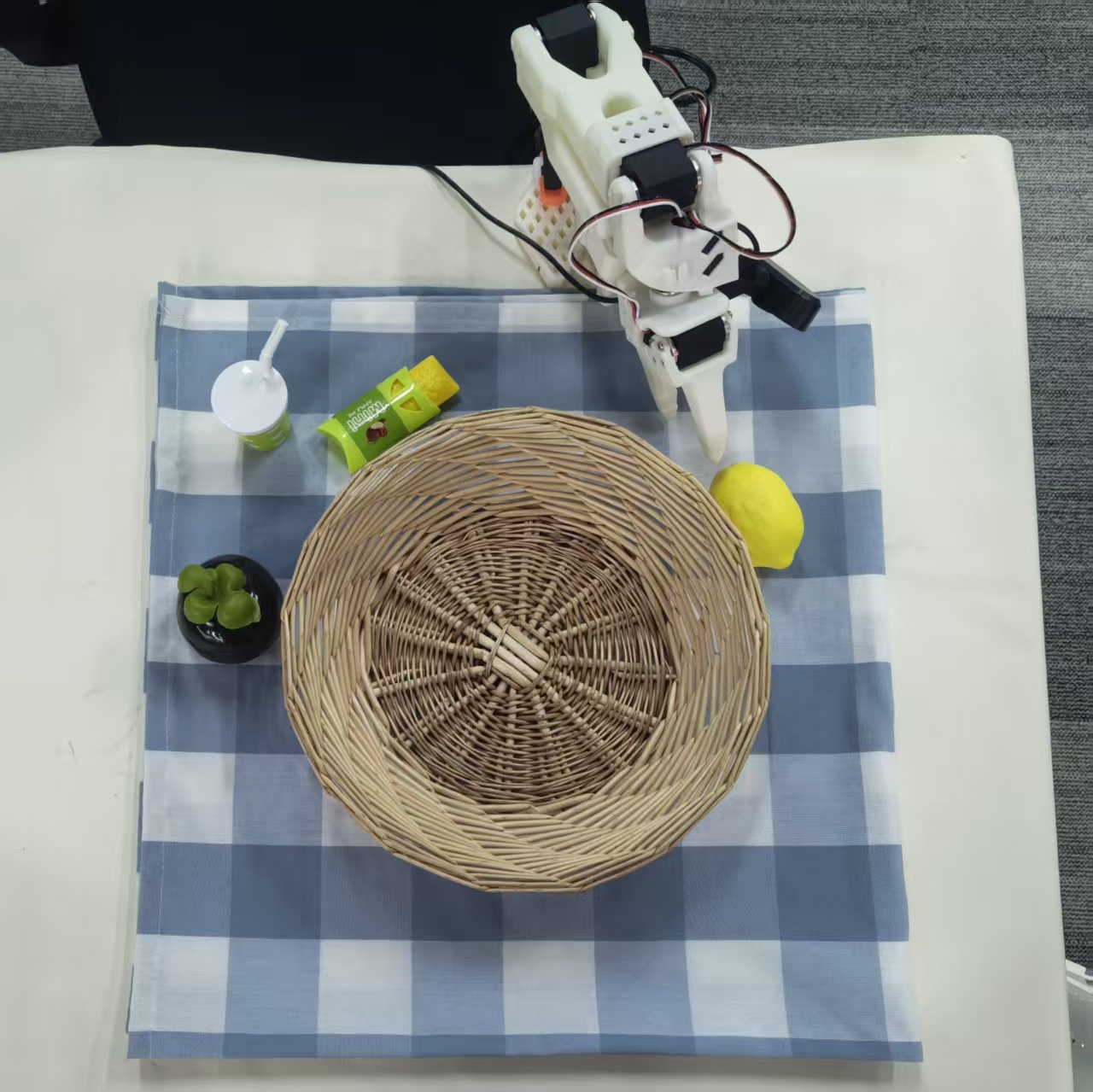}
        \caption{\begin{tabular}[t]{@{}c@{}} 
            \textbf{Train: Canonical} \\ 
            Short-horizon \\ 
            Layout 2 \\ 
            Blue tablecloth 
        \end{tabular}}
        \label{fig:train_t1_l2_b}
    \end{subfigure}
    \hfill
    \begin{subfigure}[b]{0.24\textwidth}
        \centering
        \includegraphics[width=\textwidth]{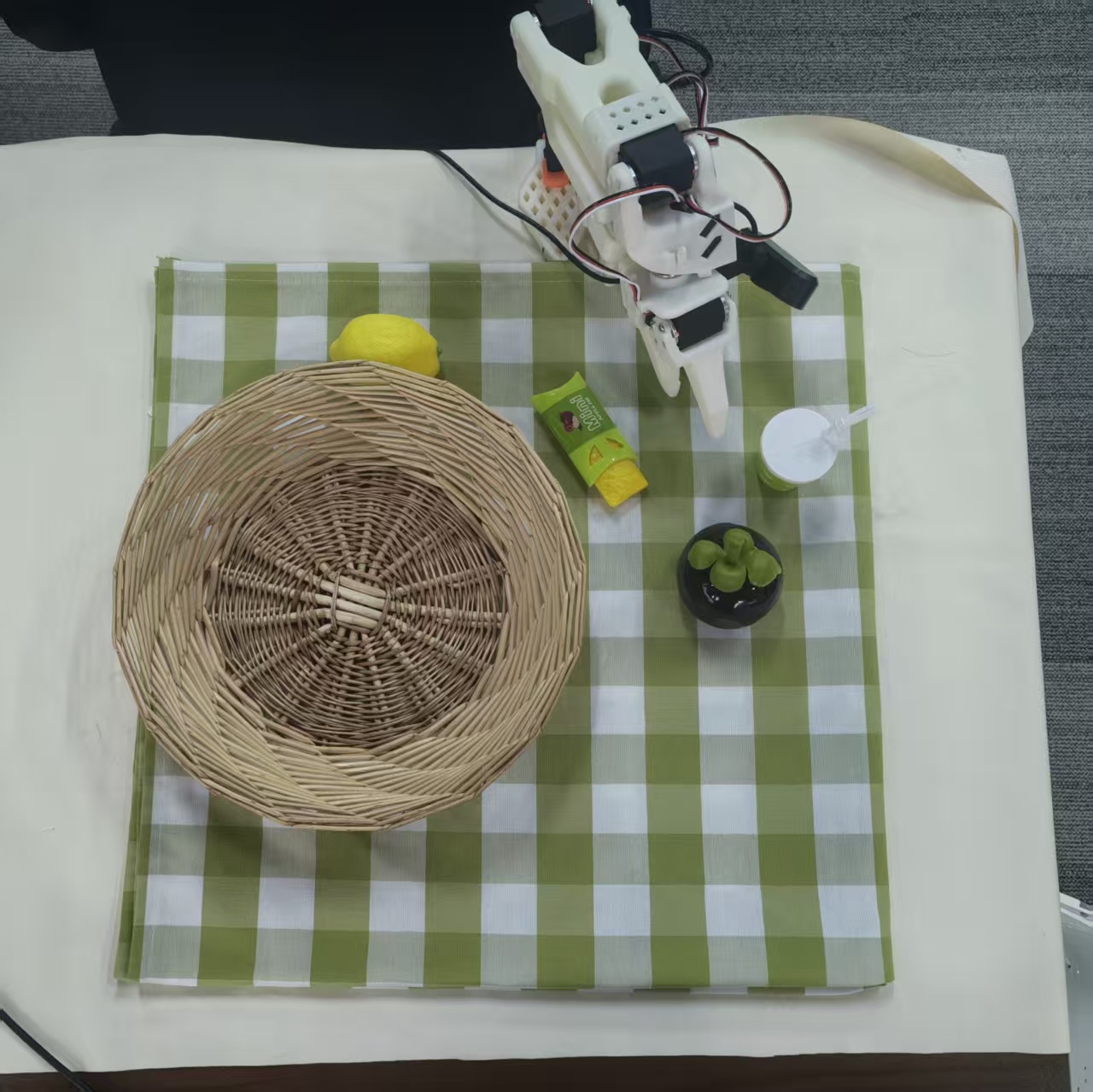}
        \caption{\begin{tabular}[t]{@{}c@{}} 
            \textbf{Train: Canonical} \\ 
            Long-horizon \\ 
            Layout 1 \\ 
            Green tablecloth 
        \end{tabular}}
        \label{fig:train_t2_l1_g}
    \end{subfigure}
    \hfill
    \begin{subfigure}[b]{0.24\textwidth}
        \centering
        \includegraphics[width=\textwidth]{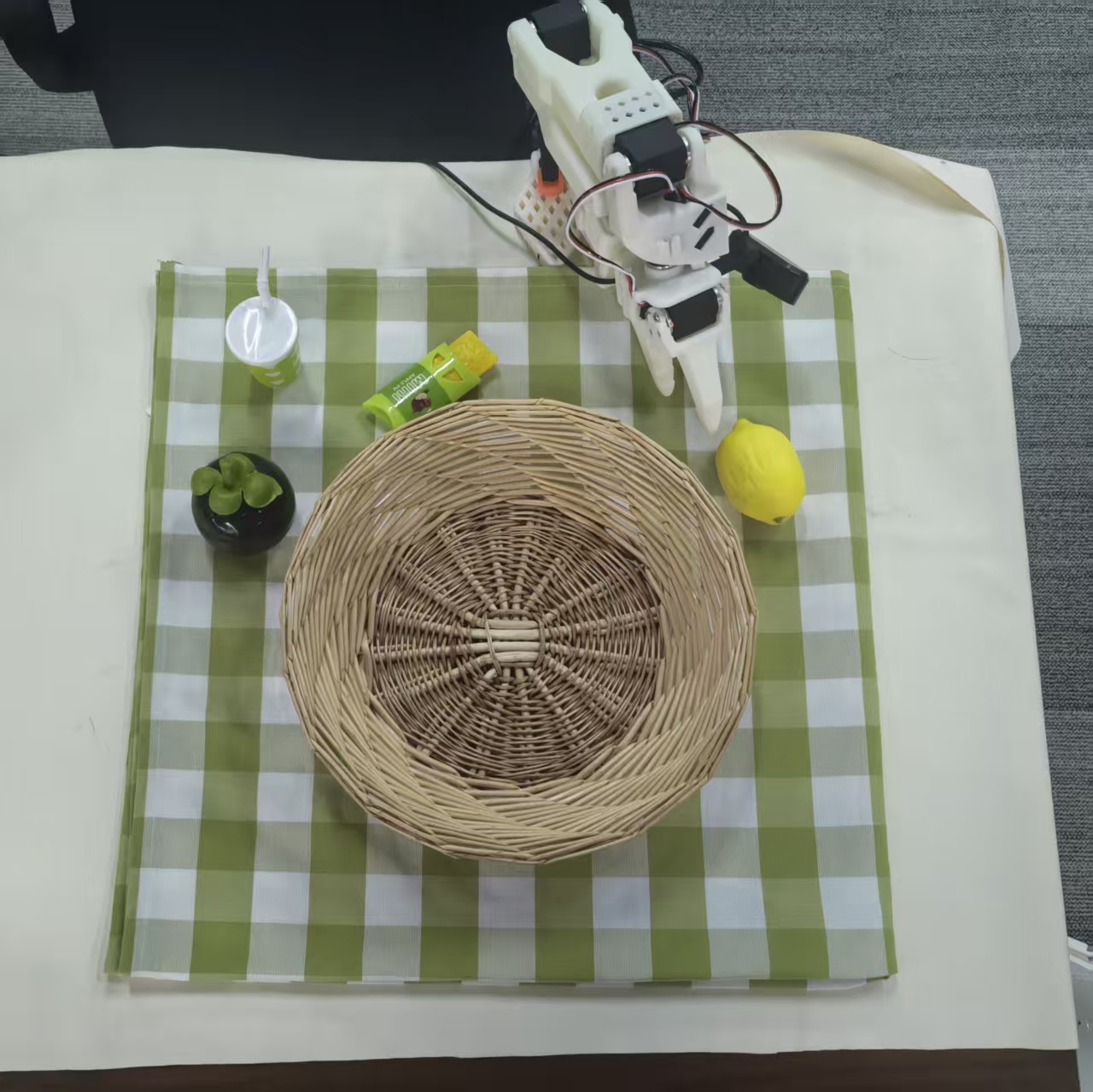}
        \caption{\begin{tabular}[t]{@{}c@{}} 
            \textbf{Train: Canonical} \\ 
            Long-horizon \\ 
            Layout 2 \\ 
            Green tablecloth 
        \end{tabular}}
        \label{fig:train_t2_l2_g}
    \end{subfigure}
    
\caption{
\textbf{Real-world evaluation protocol.}
We evaluate model robustness across two task horizons (short vs.\ long) and two spatial layouts. 
To disentangle cognitive reasoning from kinematic execution, models are fine-tuned on canonical configurations (a–d) and then subjected to targeted causal interventions (e.g., partial target absence or pre-placed objects).
These perturbations expose underlying failure modes and reveal the ``illusion of reasoning'' in physical execution.
}
    \label{fig:real_world_setup}
\end{figure*}

The simulation diagnostics (Section~\ref{sec:main_analysis}) and the mechanistic analysis (Section \ref{sec:degradation}) reveal significant reasoning deficiencies in state-of-the-art VLAs.
A key question remains: are these failures artifacts of the simulation gap, or do they reflect fundamental limitations in real-world deployment?
To answer this, we conduct targeted physical stress tests on the SO101 robotic platform.
Our results confirm that the illusion of embodied reasoning is not a simulation artifact, but a persistent phenomenon in continuous physical control.

\subsection{Experimental Setup} 

As shown in Figure~\ref{fig:real_world_setup}, the workspace consists of a fruit basket, two semantic targets (a lemon and a mangosteen), and two distractors.
We define four training variations spanning two task horizons and two spatial layouts: 1) short-horizon (``\textit{Put the lemon into the fruit basket}'') and 2) long-horizon (``\textit{Put all the fruits into the fruit basket}'').
Using $\pi_{0.5}$~\cite{intelligence2025pi05}, we collect a constrained dataset of 100 teleoperated trajectories (25 demonstrations per variation) and fine-tune the model for 10,000 steps. 
This setup ensures that the policy acquires the necessary sensorimotor capabilities ($z_a$) to execute the canonical tasks.

Despite this, our initial real-world experiments reveal that the learned policy behaves primarily as an open-loop trajectory executor rather than a reasoning agent.
As an early symptom of representation $z$ degradation, the model exhibits extreme sensitivity to object placement: relocating a target to an unseen configuration causes the policy to ignore the actual object and instead execute a grasp at memorized coordinates associated with training distractors.
Even within seen layouts, successful execution often requires objects to be positioned precisely at their canonical training locations.

We interpret this behavior as spatial overfitting --- the first stage of the ``illusion of reasoning'', where robust semantic grounding is replaced by brittle coordinate memorization.
However, this alone does not clarify whether failures are limited to perceptual grounding or extend to higher-level task reasoning.
To disentangle these factors, we enforce kinematic isolation in all subsequent diagnostics: target objects are placed within known reachable configurations.
By ensuring that low-level motor primitives remain valid, any remaining failures can be attributed to higher-level cognitive breakdowns, such as deficits in state tracking and causal reasoning, rather than execution constrains.

\subsection{The Illusion of Competence in Short-Horizon Diagnostics}
After enforcing kinematic isolation to ensure functional reachability, we first evaluate the policy using short-horizon instructions (e.g., ``\textit{Put the lemon into the fruit basket}''). 
As shown in Table~\ref{tab:short_horizon_matrix}, the policy exhibits strong apparent competence under standard configurations: it correctly grasps the target when present (Row 1) and appropriately terminates when the target is already in the receptacle (Row 2).

Failure cases, such as grasping empty space or selecting distractors, emerge only when the target is entirely absent (Row 3).
This behavior suggests a lack of true semantic grounding, with the policy reverting to memorized kinematic patterns when expected visual cues are missing.
However, treating such out-of-distribution failures as definitive evidence of reasoning breakdown is problematic, as the underlying task premise is itself violated.
This contrast exposes a critical limitation of short-horizon diagnostics: they are structurally incapable of revealing deep reasoning deficits in VLAs.
The limited execution horizon allows policies to succeed via simple reactive heuristics, such as mapping salient visual feature (e.g., color) directly to grasp primitive, without requiring causal state tracking or sequential planning. 
To expose the underlying illusion of reasoning, evaluation must instead move to long-horizon settings, where such reactive shortcuts are insufficient and genuine reasoning becomes necessary.

\begin{table}[t]
    \centering
\caption{
\textbf{Short-horizon diagnostic matrix revealing the limits of surface-level success.} 
We evaluate the task ``\textit{Put the lemon into the fruit basket}'' by varying the presence of the target and distractor. 
In standard settings (Rows 1–2), the policy achieves consistent success via reactive heuristics (e.g., color-triggered grasping and occupancy-based termination).
However, failure in Row 3 (target absent) exposes a lack of true semantic grounding, despite such cases often being dismissed as out-of-distribution in standard benchmarks.
This matrix illustrates that short-horizon tasks can be solved through superficial strategies, making them an unreliable proxy for genuine embodied reasoning.
   }
    \label{tab:short_horizon_matrix}
    \vspace{0.1cm}
    \scalebox{.9}{
    \setlength{\tabcolsep}{4pt} 
    \renewcommand{\arraystretch}{1.4} 
    
    \begin{tabular}{c | >{\centering\arraybackslash}p{4.5cm} >{\centering\arraybackslash}p{4.5cm} >{\centering\arraybackslash}p{4.5cm}}
        \toprule
        \diagbox[width=3.5cm]{\textbf{Target}}{\textbf{Distractor}} & \textbf{On Table} & \textbf{In Basket} & \textbf{Absent} \\
        \midrule
        
        \textbf{On Table} & 
        \cellcolor{green!10}\textbf{Success}\par(Accurate Grasping) & 
        \cellcolor{green!10}\textbf{Success}\par(Accurate Grasping) & 
        \cellcolor{green!10}\textbf{Success}\par(Accurate Grasping) \\
        \hline
        
        \textbf{In Basket} & 
        \cellcolor{green!10}\textbf{Correct Freeze}\par(Recognizes Done) & 
        \cellcolor{green!10}\textbf{Correct Freeze}\par(Recognizes Done) & 
        \cellcolor{green!10}\textbf{Correct Freeze}\par(Recognizes Done) \\
        \hline
        
        \textbf{Absent} & 
        \cellcolor{gray!10}\textbf{Ambiguous}\par(Grasps Distractor) & 
        \cellcolor{gray!10}\textbf{Ambiguous}\par(Freezes / Air Grasp) & 
        \cellcolor{gray!10}\textbf{Ambiguous}\par(Air Grasp) \\
        
        \bottomrule
    \end{tabular}}
\end{table}

\begin{table}[t]
    \centering
\caption{
\textbf{Long-horizon diagnostic matrix revealing sequential reasoning failures.} 
We evaluate the task ``\textit{Put all the fruits into the fruit basket}'' under partial-absence causal interventions.
Failure cases, primarily driven by proprioceptive overfitting, are highlighted in light red. The gray cell denotes an ambiguous configuration that violates the task’s underlying premise.
}
    \label{tab:long_horizon_matrix}
    \vspace{0.1cm}
    \scalebox{.9}{
    \setlength{\tabcolsep}{4pt} 
    
    \renewcommand{\arraystretch}{1.4} 
    
    \begin{tabular}{c | >{\centering\arraybackslash}p{4.5cm} >{\centering\arraybackslash}p{4.5cm} >{\centering\arraybackslash}p{4.5cm}}
        \toprule
        \diagbox[width=3.5cm]{\textbf{Lemon (T1)}}{\textbf{Mang. (T2)}} & \textbf{On Table} & \textbf{In Basket} & \textbf{Absent} \\
        \midrule
        
        \textbf{On Table} & 
        \cellcolor{green!10}\textbf{Success}\par(Packs L $\rightarrow$ M) & 
        \cellcolor{green!10}\textbf{Success}\par(Packs L $\rightarrow$ Ends) & 
        \cellcolor{red!10}\textbf{Behavioral Inertia}\par(Packs L $\rightarrow$ Air Grasp) \\
        \hline
        
        \textbf{In Basket} & 
        \cellcolor{red!10}\textbf{Phase Conflict I}\par(Freezes / Skips L) & 
        \cellcolor{green!10}\textbf{Success}\par(Recognizes Done) & 
        \cellcolor{red!10}\textbf{Phase Conflict II}\par(Oscillates / Air Grasp) \\
        \hline
        
        \textbf{Absent} & 
        \cellcolor{green!10}\textbf{Success}\par(Skips L, Packs M) & 
        \cellcolor{green!10}\textbf{Correct Freeze}\par(Recognizes Done) & 
        \cellcolor{gray!10}\textbf{Ambiguous}\par(Air Grasp) \\
        
        \bottomrule
    \end{tabular}}
\end{table}

\subsection{Deconstructing Cognitive Breakdowns in Long-Horizon Tasks}

To unmask the limitations of reactive control, we extend our evaluation to long-horizon tasks (e.g., ``\textit{Put all the fruits into the fruit basket}''). 
By introducing targeted causal intervention (Table~\ref{tab:long_horizon_matrix}), we deliberately disrupt the expected execution chain.
This setup disentangles genuine sequential reasoning from memorized heuristics, revealing two primary failure modes:

\begin{itemize}
\item \textbf{Behavioral inertia and proprioceptive overfitting.}
When the first target is present but the second is absent (Table~\ref{tab:long_horizon_matrix}, top-right), the policy correctly packs the lemon but subsequently performs an ``air grasp'' at the expected location of the missing mangosteen. Completing the first subtask induces a strong sequential prior.
Due to the degraded semantic representation, the policy fails to verify the visual scene and instead over-relies on its proprioceptive state, blindly triggering the next step of a memorized trajectory despite the absence of valid visual preconditions.

\item \textbf{Phase conflict and temporal disorientation.} 
When the lemon is pre-placed in the basket (Table~\ref{tab:long_horizon_matrix}, middle row), the visual observation indicates an intermediate task state (Step 1 completed), while the robot’s proprioceptive state (resting at the home position) corresponds to $t=0$. 
This conflicting conditioning leads to two characteristic failure patterns.
First, if the second target is present, the mismatch often results in near-zero velocity outputs, causing the policy to ``freeze'' instead of transition smoothly to the next sub-task. 
Second, if the second target is also absent, the policy exhibits erratic exploratory motions ---extending the arm without a valid target before returning to rest. 
In both cases, the policy fails to perform causal state tracking based on visual feedback, relying instead on rigid assumptions tied to initial conditions.
\end{itemize}

These real-world diagnostics provide a mechanistic closure to the failures identified throughout this work.
Crucially, they directly address the central question posed at the outset: the observed reasoning deficits are not artifacts of the sim-to-real gap, but reflect a fundamental limitation of current VLA paradigms.
When the semantic representation $z$ degrades due to architectural bottlenecks, the policy is forced to overfit to sensorimotor representations $z_a$, leading to brittle heuristics that collapse under causal interventions.
These findings underscore the need for next-generation architectures that can preserve high-level semantic reasoning while maintaining the efficiency required for high-frequency control.

\section{Conclusion}
\label{sec:conclusions}

In this work, we unmask the ``illusion of embodied reasoning'' in state-of-the-art VLAs using BeTTER, a diagnostic benchmark that explicitly isolates high-level cognitive breakdowns from low-level motor execution limits. 
We demonstrate that impressive benchmark performance often masks shortcut learning, as models exhibit severe semantic feature collapse, behavioral inertia, and causal state tracking failures under controlled causal interventions.
Crucially, we trace these vulnerabilities to structural architectural bottlenecks—such as capacity compression and myopic perceptual downsampling—which systematically degrade the model's foundational semantic representation ($z$). 
Our comparative analysis reveals that highly static evaluation environments effectively mask this degradation by allowing policies to max out success rates through overfitting to sensorimotor representation ($z_a$). However, under out-of-distribution shifts, this brittle trajectory memorization catastrophically fails, proving that open-world generalization strictly requires genuine semantic grounding.
Supported by rigorous real-world validations, our findings provide a mechanistic lens to distinguish genuine physical intelligence from superficial behavioral competence, highlighting the critical need for future VLA paradigms to resolve the structural tension between high-frequency control and high-level semantic grounding.

\clearpage

\bibliographystyle{unsrt}
\bibliography{ref}

\clearpage

\beginappendix


\section{Data Recipes, Evaluation Framework, and Implementation Details}

\subsection{Pre-training Data Recipe}
\label{sec:data_recipe}

To systematically train and evaluate the capabilities of our models, we curate a large-scale mixture of Vision-Language Model (VLM) datasets and Vision-Language-Action (VLA) datasets. The VLM data is explicitly designed to preserve and enhance the foundational semantic representation ($z$)---including general visual-text understanding, multi-frame reasoning, and dense spatial grounding. Conversely, the VLA data is derived from the Open X-Embodiment (OXE) registry~\cite{o2024open} to establish the sensorimotor and kinematic prior ($z_a$) required for continuous control. The detailed statistics are presented in Table~\ref{tab:vlm_data} and Table~\ref{tab:vla_data}. The specific mixing weights and sampling strategies used during co-training are detailed in Section~\ref{sec:supp_mixture}.

\begin{table}[htbp]
    \centering
    \caption{\textbf{Summary of Vision-Language Model (VLM) Datasets.} These diverse datasets are utilized during the pre-training phase to preserve the model's high-level cognitive capabilities, acting as a representational regularizer for general visual-text understanding, multi-frame reasoning, and dense spatial grounding.}
    \label{tab:vlm_data}
    \vspace{0.1cm}
    \scalebox{0.9}{
        \setlength{\tabcolsep}{4pt}
        \renewcommand{\arraystretch}{1.3}
        \begin{tabular}{>{\raggedright\arraybackslash}p{4.6cm} p{9.3cm} c r}
            \toprule
            \textbf{Cognitive Category} & \textbf{Dataset Sources} & \textbf{Modality} & \textbf{Samples} \\
            \midrule
            \textbf{General SFT} & LLaVA v1.5 (158k, 665k)~\cite{liu2024improved}, LLaVA OV SI~\cite{li2024llava-ov} & Img-Txt & $\sim$4.76M \\
            \textbf{Video-Text SFT} & LLaVA Video (178k)~\cite{zhang2024llava-video} & Vid-Txt & $\sim$1.39M \\
            \textbf{2D Spatial / Grounding} & RefCOCO~\cite{yu2016refcoco}, RoboPoint~\cite{yuan2024robopoint}, RoboRefIt~\cite{lu2023vl}, ShareRobot~\cite{ji2025robobrain}, All-Seeing v2 (Rec/Mix)~\cite{wang2024asv2}, EO-1.5M QA~\cite{qu2025eo1}, Pixmo Points~\cite{deitke2025molmo-pixmo} & Img-Txt & $\sim$7.47M \\
            \textbf{Multi-frame Planning} & ShareRobot Planning Suite~\cite{ji2025robobrain} & Vid-Txt & $\sim$1.02M \\
            \midrule
            \textbf{Total VLM Data} & & & \textbf{$\sim$14.64M} \\
            \bottomrule
        \end{tabular}
    }
\end{table}

\begin{table}[htbp]
    \centering
    \caption{\textbf{Summary of Open X-Embodiment (VLA) Datasets.} This large-scale subset of the OXE registry is utilized for continuous control pre-training. It provides the dense, high-frequency manipulation trajectories necessary to establish the robotic sensorimotor priors.}
    \label{tab:vla_data}
    \vspace{0.1cm}
    \scalebox{0.95}{
    \setlength{\tabcolsep}{16pt}
    \renewcommand{\arraystretch}{1.2}
    \begin{tabular}{l r r}
        \toprule
        \textbf{Dataset Name} & \textbf{Episodes} & \textbf{Frames}\\
        \midrule
        DROID~\cite{khazatsky2024droid} & 92,233 & 27,044,326\\
        Language Table~\cite{lynch2023interactive} &  442,226 & 7,045,476 \\
        BC-Z~\cite{jang2021bc} & 39,350 & 5,471,693 \\
        Fractal~\cite{brohan2022rt} & 87,212 & 3,786,400 \\
        KUKA~\cite{kalashnikov2018qt} & 209,880 & 2,455,879 \\
        Bridge Orig~\cite{walke2023bridgedata} & 53,192 & 1,893,026 \\
        \midrule
        \textbf{Total VLA Data} & \textbf{924,093} & \textbf{$\sim$47.69M} \\
        \bottomrule
    \end{tabular}
    }
\end{table}

\subsection{Pre-training Mixture Strategy and Planning Objective}
\label{sec:supp_mixture}

Following the data curation detailed in Section \ref{sec:data_recipe}, balancing continuous control proficiency with the retention of spatial-semantic knowledge is critical during the pre-training phase. 

\vspace{1mm}
\noindent\textbf{Data Sampling Ratio.} \\
For the \textit{VLA + VLM} co-training configuration, we employ a strategically skewed sampling distribution to prioritize motor execution while maintaining the VLM data as a representational regularizer. Specifically, the empirical sampling weight is set to 8.0 for the Open X-Embodiment (VLA) mixture, and 0.5 for each of the four VLM subsets (General SFT, Video-Text, 2D Spatial, and Multi-frame Planning).

\vspace{1mm}
\noindent\textbf{Hardware and Compute Scaling Strategy.} \\
To ensure equitable model convergence and fair comparison across pre-training recipes of drastically different scales, we scale our computational resources proportionally to the dataset sizes. All configurations are trained using NVIDIA A800 GPUs. Specifically, \textit{VLA-Only} configuration is trained for 100k steps across 16 GPUs, while the full \textit{VLA + VLM Co-training} configuration utilizes 24 GPUs for 100k steps. This resource allocation strategy ensures that none of the model configurations underfit their respective data mixtures.

\subsection{Standardized VLM Evaluation Framework}
\label{sec:supp_vlm_eval}

To ensure a rigorous and fair evaluation of foundational capabilities—rather than instruction-following strictness—we constructed a standardized evaluation framework for the VLM benchmarks (EmbSpatial~\cite{du2024embspatial}, RefSpatial~\cite{zhou2025roborefer}, and EgoPlan-Bench2~\cite{qiu2024egoplan}).

For models pre-trained in our pipeline, we explicitly aligned the evaluation prompts with their respective training templates to accurately measure the acquired knowledge. Conversely, for the baseline InternVL models, since their exact pre-training formatting distributions are opaque, we employed standard, zero-shot prompts. 
Crucially, to prevent these baselines from being artificially penalized for formatting deviations, we integrated an LLM (GPT-4o-mini) as an unbiased answer extractor. For instance, if a baseline model correctly outputs spatial coordinates but surrounds them with unexpected explanation, the LLM robustly isolates the exact values for standardized scoring. This strict post-processing decouples genuine semantic and spatial reasoning capabilities from formatting constraints.

\subsection{Fine-Tuning on the BeTTER}
\label{sec:supp_better_ft}

To adapt the evaluated VLAs ($\pi_{0.5}$~\cite{intelligence2025pi05}, GR00T-N1.6~\cite{nvidia2025gr00t}, and Being-H0.5~\cite{luo2026beingh05}) to the BeTTER tasks, we standardize the training hyperparameters to ensure fair comparison. All models are fine-tuned for $30,000$ steps with an effective global batch size of $128$ (achieved natively for $\pi_{0.5}$ and GR00T-N1.6, and via dynamic sequence packing for Being-H0.5).

A critical consideration during BeTTER fine-tuning is the preservation of pre-trained reasoning capabilities. For GR00T-N1.6, our empirical tests revealed that full-parameter updates on the limited BeTTER dataset led to severe catastrophic forgetting of its semantic priors. Because BeTTER explicitly diagnoses foundational reasoning rather than domain adaptation, we adhere to the official recommendations for GR00T-N1.6: we freeze the Vision Encoder and LLM backbone, focusing updates exclusively on the projectors and action policy heads.

\subsection{Evaluation on LIBERO and CALVIN Benchmarks}
\label{sec:supp_libero_calvin}

For the downstream generalization evaluations presented in Section \ref{sec:sterile_bubble_and_validation}, we maintain the architectural setup, batch size, learning rate, and all other key hyperparameters identical to those used in the BeTTER evaluation. The primary differences lie in the training duration, dataset composition, and specific unfreezing strategies to handle domain shifts:

\begin{itemize}
    \item \textbf{LIBERO Benchmark \cite{liu2023libero}:} We mix the training data from all four task suites (LIBERO-Spatial, Object, Goal, and 10-Task) and fine-tune the models for a total of $60,000$ steps.
    
    \item \textbf{CALVIN Benchmark \cite{mees2022calvin}:} We utilize the standard ABC$\rightarrow$D split and fine-tune the models for $50,000$ steps. Conversely to the BeTTER setup, our empirical validation on CALVIN demonstrated that full-parameter tuning achieves more competitive results for all three models. This empirically driven adaptation yields significantly better performance on the challenging long-horizon sequential tasks.
\end{itemize}

\newpage
\section{The BeTTER Benchmark: Detailed Designs}

This appendix provides supplementary technical details for the programmatic pipelines introduced in Section \ref{sec:asset_and_task}. Below, we present the complete prompt templates, additional qualitative examples of our asset retrieval process, and details regarding the privileged logging mechanisms.

\subsection{Prompt Templates for Task Instantiation}
\label{sec:supp_prompt}

As discussed in Section \ref{sec:asset_and_task}, we utilize a structured system prompt to procedurally generate physically plausible task variations. The complete template provided to the VLM is detailed below.

\begin{tcolorbox}[
    breakable, 
    enhanced,
    colback=gray!5, 
    colframe=gray!40, 
    title=\textbf{System Prompt for VLM Task Instantiation},
    fonttitle=\bfseries\scriptsize, 
    boxsep=1pt, left=1.5pt, right=1.5pt, top=1pt, bottom=1pt, 
    arc=1mm
]
\begin{lstlisting}[
    basicstyle=\scriptsize\ttfamily\linespread{0.85}, % 核心：缩小字号并压缩行距
    breaklines=true,            
    columns=fullflexible,       
    keepspaces=true,            
    frame=none,                 
    aboveskip=0pt,              
    belowskip=0pt               
]
**Role:**
You are an expert **Procedural Scene Designer** for a robotics simulation benchmark.
Your goal is to instantiate abstract "Task Templates" into concrete, realistic, and physically plausible scenarios.

**Input:**
You will receive a **Task Template** JSON containing:
1. `meta`: Task name and task description.
2. `object_groups`: A list of abstract object groups. Each group has:
   - `group_id`: Unique identifier.
   - `description`: What this group represents.
   - `allowed_types`: Semantic categories allowed.
   - `count_range`: [min, max] number of objects to generate.

**Your Task:**
Generate **{request.num_scenarios}** distinct scenarios of this template.
For each variation:
1. **Invent a Scenario:** A specific real-world context (e.g., "Making a BLT sandwich" for a stacking task).
2. **Generate Instruction:** Write a clear, concise natural language instruction for the robot based on the `task_description` and your invented scenario, and avoid chaining all subtasks or steps to complete the task (e.g., "Make a ham sandwich on the plate.").
3. **Instantiate Objects:** For each `object_group`, select specific objects that fit the scenario and the group's constraints.
   - **Count:** You MUST generate a number of items within the `count_range`.
   - **Consistency:** The objects must make sense together (e.g., bread, lettuce, tomato).
   - **Physicality:** Ensure objects are physically suitable (e.g., don't stack a bowling ball on an egg).
   - **Properties:** Estimate the physical properties of the object:
     - `estimated_mass`: A range [min, max] in kg (e.g., [0.05, 0.1] for a slice of bread).
     - `estimated_size`: A range [min, max] in meters for the object's largest dimension (e.g., [0.1, 0.15] for a plate diameter).

**Output Format:**
You must output a **SINGLE JSON LIST** of variation objects. Do not include markdown formatting.

Schema:
[
  {{
    "scenario_name": "String (e.g., 'Making a Ham Sandwich')",
    "instruction": "String (e.g., 'Make a ham sandwich on the plate.')",
    "scene_context": "String (e.g., 'Kitchen')",
    "instantiation": {{
      "group_id_1": [
        {{
          "description": "Visual description (e.g., 'A slice of whole wheat bread')",
          "asset_query": "Search keyword for asset database (e.g., 'bread slice')",
          "instance_name": "Unique name (e.g., 'bread_bottom')",
          "estimated_mass": [0.05, 0.1],
          "estimated_size": [0.1, 0.15],
          "tags": []
        }},
        ... (repeat for N items, where min <= N <= max)
      ],
      "group_id_2": [ ... ]
    }}
  }},
  ...
]   
\end{lstlisting}
\end{tcolorbox}

\subsection{Open-Vocabulary Asset Retrieval Examples}
\label{sec:supp_asset_retrieval}

Expanding on the open-vocabulary asset integration pipeline detailed in Section \ref{sec:asset_and_task}, Table \ref{tab:asset_retrieval} provides additional qualitative examples of our one-to-many retrieval process.

\begin{table}[h]
    \centering
    \caption{\textbf{Qualitative examples of one-to-many open-vocabulary asset retrieval.} For each VLM-generated semantic description, our pipeline retrieves and filters multiple diverse 3D assets from Objaverse \cite{deitke2023objaverse}. This achieves high visual and geometric diversity across task variations. Objaverse UIDs are listed below the object images.}
    \label{tab:asset_retrieval}
    \vspace{0.2cm}
    \scalebox{.95}{
    \renewcommand{\arraystretch}{1.5}
    
    \begin{tabular}{>{\raggedright\arraybackslash}m{5.2cm} m{8.2cm}}
        \toprule
        \textbf{VLM Generated Description} & \multicolumn{1}{c}{\textbf{Retrieved 3D Assets \& Objaverse UIDs}} \\
        \midrule
        
        ``A shiny red apple'' & 
        \begin{minipage}[c]{8.2cm}
            \centering
            \vspace{0.15cm}
            \includegraphics[width=2.4cm]{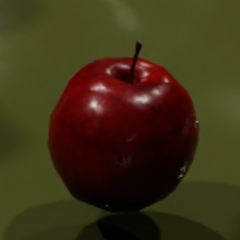} \hfill
            \includegraphics[width=2.4cm]{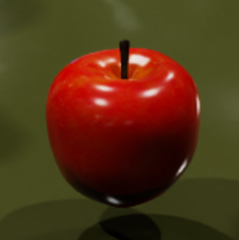} \hfill
            \includegraphics[width=2.4cm]{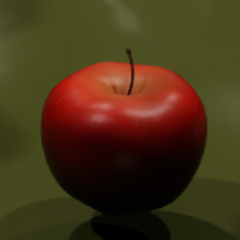} \\
            \vspace{0.1cm}
            {\centering \tiny \textit{UIDs: \\ a2056ccc0ed14427a42610e56fb7bd5f, \\ 2edb597e31b343ea9d59d79e9edbfa3c, \\ a4ecfbeef9c446c88d69fef8f5c2ff2b} \par}
            \vspace{0.15cm}
        \end{minipage} \\
        \midrule
        
        ``A blue ceramic coffee mug'' & 
        \begin{minipage}[c]{8.2cm}
            \centering
            \vspace{0.15cm}
            \includegraphics[width=2.4cm]{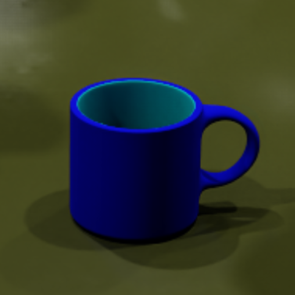} \hfill
            \includegraphics[width=2.4cm]{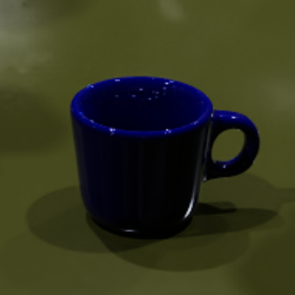} \hfill
            \includegraphics[width=2.4cm]{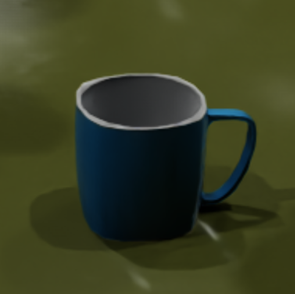} \\
            \vspace{0.1cm}
            {\centering \tiny \textit{UIDs: \\ fdbfecddc16047beb5ec47094b9c03a3, \\ dd2f219eed5a401d8c0945a4be8c6d33, \\ 2bb1d7ddbb90462ea9b27631e9376683} \par}
            \vspace{0.15cm}
        \end{minipage} \\
        \midrule
 
        ``A full, unopened can of soda'' & 
        \begin{minipage}[c]{8.2cm}
            \centering
            \vspace{0.15cm}
            \includegraphics[width=2.4cm]{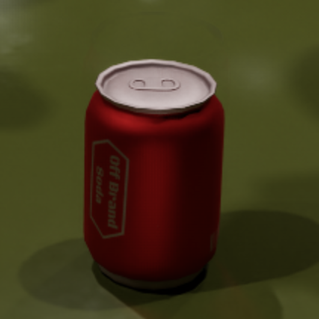} \hfill
            \includegraphics[width=2.4cm]{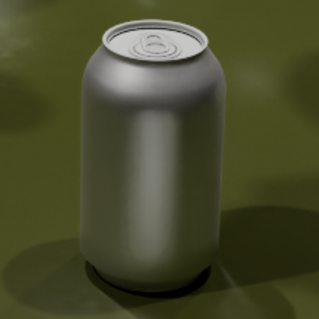} \hfill
            \includegraphics[width=2.4cm]{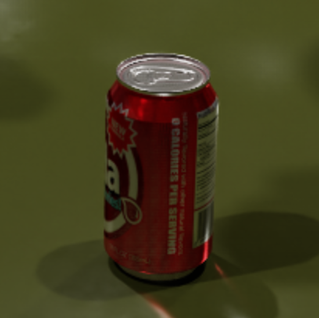} \\
            \vspace{0.1cm}
            {\centering \tiny \textit{UIDs: \\ c9c6bae3e2ec401aad393a47cc127d2a, \\ 10a464c915d0483ca58eff1514faa570, \\ e73064e272d6432cb6f1101de7f03b79} \par}
            \vspace{0.15cm}
        \end{minipage} \\
        
        \bottomrule
    \end{tabular}}
\end{table}

\subsection{VQA Generation Pipeline}
\label{sec:supp_logging_vqa}

\begin{figure}[h]
    \centering
    
    \begin{minipage}[t]{0.31\linewidth}
        \centering
        \includegraphics[width=\linewidth]{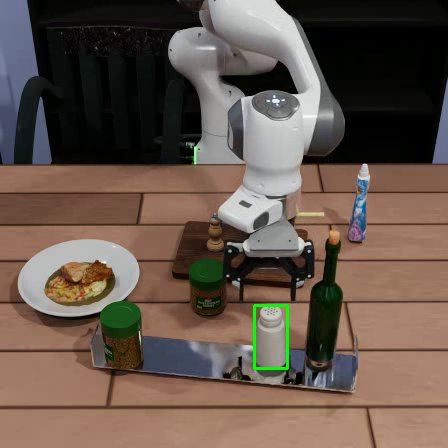}
        \vspace{0.15cm}
        
        \raggedright
        \small\textbf{1. Object Grounding} \\
        \vspace{0.05cm}
        \scriptsize
        \textit{\textbf{Q:}} \textit{What is the bounding box of the salt shaker?} \\
        \textit{\textbf{A:}} \textit{[254.0, 305.0, 287.0, 368.0]}
    \end{minipage}\hfill
    %
    \begin{minipage}[t]{0.31\linewidth}
        \centering
        \includegraphics[width=\linewidth]{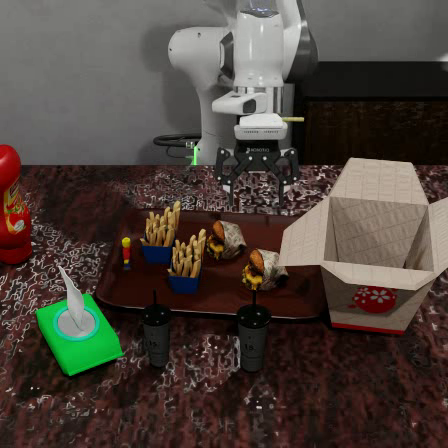}
        \vspace{0.15cm}
        
        \raggedright
        \small\textbf{2. Object Counting} \\
        \vspace{0.05cm}
        \scriptsize
        \textit{\textbf{Q:}} \textit{How many cheeseburger objects can you see?} \\
        \textit{\textbf{A:}} \textit{2}
    \end{minipage}\hfill
    %
    \begin{minipage}[t]{0.31\linewidth}
        \centering
        \includegraphics[width=\linewidth]{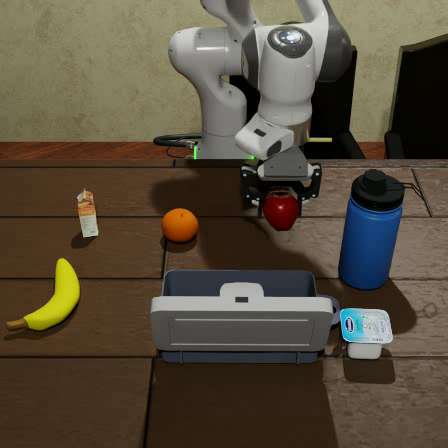}
        \vspace{0.15cm}
        
        \raggedright
        \small\textbf{3. Execution Tracking} \\
        \vspace{0.05cm}
        \scriptsize
        \textit{\textbf{Q:}} \textit{What is the robot doing?} \\
        \textit{\textbf{A:}} \textit{The robot is packing apple into the lunchbox container.}
    \end{minipage}
    
    \vspace{0.2cm}
    \caption{\textbf{Representative Examples of State-Grounded VQA.} By leveraging deterministic privileged simulation states (e.g., exact 3D coordinates, object counts, and procedural task milestones), our automated pipeline generates dense, hallucination-free visual-language supervision. The three panels demonstrate fine-grained spatial grounding, robust instance counting, and dynamic execution tracking.}
    \label{fig:vqa_samples}
\end{figure}

As discussed in Section \ref{sec:data_collection}, we anchor our VQA generation pipeline to deterministic privileged simulation states to eliminate spatial hallucinations. Figure~\ref{fig:vqa_samples} illustrates representative examples of this state-grounded VQA data. By coupling visual observations with privileged states, we generate strictly accurate QA pairs covering spatial geometry, semantic quantities, and temporal execution.

\subsection{Overview of Evaluated Tasks}
\label{sec:supp_evaluated_tasks}
To rigorously evaluate generalist physical intelligence, \benchmark comprises 10 distinct base tasks. Rather than reporting exhaustive, fine-grained scores across all combinatorial perturbations, we provide a macroscopic taxonomy of the core tasks and their physical constraints in Table~\ref{tab:task_taxonomy}. This systematic design ensures diverse physical manipulation challenges, serving as the robust foundation for our out-of-distribution dynamic perturbations.

\begin{table}[h]
    \centering
    \caption{\textbf{Template-Based Taxonomy of Base Tasks in \benchmark.} The benchmark tasks are instantiated across seven distinct constraint templates. This systematic design ensures diverse physical manipulation challenges (ranging from loose containment to precision insertion), serving as the robust foundation for our out-of-distribution causal interventions.}
    \label{tab:task_taxonomy}
    \vspace{0.1cm}
    \scalebox{0.92}{
    \renewcommand{\arraystretch}{1.5}
    \begin{tabular}{p{4.5cm} p{5.6cm} p{6.5cm}}
        \toprule
        \textbf{Constraint Template} & \textbf{Task Name} & \textbf{Core Challenge} \\
        \midrule
        
        \multirow{2}{*}{\textbf{Patterned Arrangement}} 
        & Preparing Desk for Remote Meeting & Precise relative spatial arrangement and orientation \\
        & Setting the Table for Breakfast & Semantic target selection among dense objects and relational positioning \\
        \midrule
        
        \multirow{3}{*}{\textbf{Loose Packing}} 
        & Packing a Fruit Lunch & Free-form spatial containment and target tracking \\
        & Packing a Fast Food Order & Multi-instance causal tracking (e.g., handling identical items) to prevent repetitive or omitted actions \\
        & Tidying up a Desk & Distractor-robust clearing \\
        \midrule

        \textbf{Constrained Positioning} 
        & Preparing Morning Coffee & High-precision alignment with static environmental fixtures \\
        \midrule
        
        \textbf{Container Loading} 
        & Organize Office Drawer & Understanding articulated physical constraints (e.g., closing obstructing drawers) and spatial insertion \\
        \midrule

        \textbf{Logical Assembly} 
        & Assembling a Cheeseburger & Strict semantic ordering and multi-layer structural balance \\        
        \midrule
        
        \textbf{Precision Insertion} 
        & Organizing Seasoning Rack & Tight-clearance spatial insertion and strict collision avoidance \\        
        \midrule

        \textbf{Recursive Stacking} 
        & Stacking Plates for Dinner & Iterative geometric alignment and balance maintenance \\        
        
        \bottomrule
    \end{tabular}
    }
\end{table}

\end{document}